\documentclass{article}

\usepackage{arxiv}

\usepackage[utf8]{inputenc} 
\usepackage[T1]{fontenc}    
\usepackage{hyperref}       
\usepackage{url}            
\usepackage{booktabs}       
\usepackage{amsfonts}       
\usepackage{nicefrac}       
\usepackage{microtype}      
\usepackage{lipsum}		
\usepackage{graphicx}
\usepackage{natbib}
\usepackage{doi}

\usepackage{amsthm}
\newtheorem{example}{Example}
\newtheorem{definition}{Definition}
\newtheorem{proposition}{Proposition}

\usepackage{amsmath,amssymb}
 
\DeclareMathOperator*{\argmax}{argmax}
\usepackage[algo2e]{algorithm2e} 
\usepackage{mathtools} 
\usepackage{booktabs} 
\usepackage{tikz} 
\usepackage{subcaption}
\usepackage{siunitx} 
\usepackage{multirow}
\usepackage{tabularx}
\usepackage{colortbl}
\usepackage{amsthm}

\usepackage{caption}
\usepackage{multicol}
\newcommand{\eat}[1]{}

\newcommand{\ourfig}{SSMP\xspace}

\newcommand{\our}{\texttt{SSMP}\xspace}
\newcommand{\spn}{\texttt{Max}\xspace}
\newcommand{\mle}{\texttt{ML}\xspace}
\newcommand{\seq}{\texttt{Seq}\xspace}

\newcommand{\mpe}{MPE\xspace}
\newcommand{\mmap}{MMAP\xspace}

\usepackage{tikz}
\newcommand{\myscale}{0.02}
\usetikzlibrary{positioning,quotes}

\date{} 					

\title{Neural Network Approximators for Marginal MAP in Probabilistic Circuits}

\author{Shivvrat Arya \\
	Department of Computer Science \\
	The University of Texas at Dallas\\
	shivvrat.arya@utdallas.edu 
	\And
	Tahrima Rahman \\
	Department of Computer Science\\
	The University of Texas at Dallas\\
	tahrima.rahman@utdallas.edu 
        \And
        Vibhav Gogate \\
	Department of Computer Science\\
	The University of Texas at Dallas\\
	vibhav.gogate@utdallas.edu 
}

\renewcommand{\shorttitle}{Neural Network Approximators for Marginal MAP in Probabilistic Circuits}

\hypersetup{
pdftitle={Neural Network Approximators for Marginal MAP in Probabilistic Circuits},
pdfsubject={cs.AI, cs.LG},
pdfauthor={Shivvrat Arya, Tahrima Rahman, Vibhav Gogate},
pdfkeywords={Probabilistic Circuits, Graphical Models, Probabilistic Inference},
}

\begin{document}
\maketitle

\begin{abstract}
 Probabilistic circuits (PCs) such as sum-product networks efficiently represent large multi-variate probability distributions. They are preferred in practice over other probabilistic representations, such as Bayesian and Markov networks, because PCs can solve marginal inference (MAR) tasks in time that scales linearly in the size of the network. Unfortunately, the most probable explanation (MPE) task and its generalization, the marginal maximum-a-posteriori (MMAP) inference task remain NP-hard in these models. Inspired by the recent work on using neural networks for generating near-optimal solutions to optimization problems such as integer linear programming, we propose an approach that uses neural networks to approximate MMAP inference in PCs. The key idea in our approach is to approximate the cost of an assignment to the query variables using a continuous multilinear function and then use the latter as a loss function. The two main benefits of our new method are that it is self-supervised, and after the neural network is learned, it requires only linear time to output a solution. We evaluate our new approach on several benchmark datasets and show that it outperforms three competing linear time approximations: max-product inference, max-marginal inference, and sequential estimation, which are used in practice to solve MMAP tasks in PCs.
\end{abstract}


\section{Introduction}\label{sec:intro}
Probabilistic circuits (PCs) \cite{choiProbCirc20} such as sum-product networks (SPNs) \cite{poon2011sum}, arithmetic circuits \cite{darwiche2003differential}, AND/OR graphs \cite{dechter2007and}, cutset networks \cite{rahman2014cutset}, and probabilistic sentential decision diagrams \cite{kisa2014probabilistic}  represent a class of \textit{tractable probabilistic models} which are often used in practice to compactly encode a large multi-dimensional joint probability distribution. Even though all of these models admit linear time computation of marginal probabilities (MAR task), only some of them \cite{NEURIPS2021_vergari,peharz2015foundations}, specifically those without any latent variables or having specific structural properties, e.g., cutset networks, selective SPNs \cite{peharz2016latent}, AND/OR graphs having small contexts, etc., admit tractable most probable explanation (MPE) inference\footnote{The MPE inference task is also called full maximum-a-posteriori (full MAP) inference in literature. In this paper, we adopt the convention of calling it MPE.}. 

However, none of these expressive PCs can efficiently solve the \textit{marginal maximum-a-posteriori (MMAP) task} \cite{peharz2015foundations,NEURIPS2021_vergari}, a task that combines MAR and MPE inference. 
More specifically, the distinction between MPE and MMAP tasks is that, given observations over a subset of variables (evidence), the MPE task aims to find the most likely assignment to all the non-evidence variables. In contrast, in the MMAP task, the goal is to find the most likely assignment to a subset of non-evidence variables known as the query variables, while marginalizing out non-evidence variables that are not part of the query.
The MMAP problem has numerous real-world applications, especially in health care, natural language processing, computer vision, linkage analysis and diagnosis where hidden variables are present and need to be marginalized out \cite{bioucasdiasBayesianImageSegmentationusinghiddenfields2016, kiselevPOMDPPlanningMarginalMAPProbabilisticInferenceGenerative, leeApplyingMarginalMapSearchProbabilisticConformantb,NIPS2015_faacbcd5}. 

In terms of computational complexity, both MPE and MMAP tasks are at least NP-hard in SPNs, a popular class of PCs \cite{peharz2015foundations,conaty17}. Moreover, it is also NP-hard to approximate MMAP in SPNs to $2^{n^\delta}$ for fixed $0 \leq \delta < 1$,
where $n$ is the input size \cite{conaty17,mei2018maximum}. It is also known that the MMAP task is much harder than the MPE task and is NP-hard even on models such as cutset networks and AND/OR graphs that admit linear time MPE inference \cite{park&darwiche04,decamposNewComplexityResultsMAPBayesianNetworks}.

To date, both exact and approximate methods have been proposed in literature for solving the MMAP task in PCs. Notable exact methods include branch-and-bound search \cite{mei2018maximum}, reformulation approaches which encode the MMAP task as other combinatorial optimization problems with widely available solvers \cite{maua2020two} and circuit transformation and pruning techniques \cite{choi2022solving}. These methods can be quite slow in practice and are not applicable when fast, real-time inference is desired. As a result, approximate approaches that require only a few passes over the PC are often used in practice. A popular approximate approach is to compute an MPE solution over both the query and unobserved variables and then project the MPE solution over the query variables \cite{poon2011sum,rahman2019look}. Although this approach can provide fast answers at query time, it often yields MMAP solutions that are far from optimal. 

In this paper, we propose to address the limitations of existing approximate methods for MMAP inference in PCs by using neural networks (NNs), leveraging recent work in the \emph{learning to optimize} literature \cite{li2016learning,fioretto_optimalpowerflows_2020,  donti_methodoptimizationhard_2020, zamzam_solutionsextremelyfast_2020, park_learningconstrainedoptimization_2023}.
In particular, several recent works have shown promising results in using NNs to solve both constrained and unconstrained optimization problems (see \citet{park_learningconstrainedoptimization_2023} and the references therein). 

The high-level idea in these works is the following: given data, train NNs, either in a supervised or self-supervised manner, and then use them at test time to predict high-quality, near-optimal solutions to future optimization problems. A number of reasons have motivated this idea of \textit{learning to optimize} using NNs: 1) NNs are good at approximating complex functions (distributions), 2) once trained, they can be faster at answering queries than search-based approaches, and 3) with ample data, NNs can learn accurate mappings of inputs to corresponding outputs. This has led researchers to employ NNs to approximately answer probabilistic inference queries such as MAR and MPE in Bayesian and Markov networks \cite{yoon2019inference, cui_variational_2022}. To the best of our knowledge, there is no prior work on solving MMAP in BNs, MNs, or PCs using NNs.  

This paper makes the following contributions. First, we propose to learn a neural network (NN) approximator for solving the MMAP task in PCs. Second, by leveraging the tractability of PCs, we devise a loss function whose gradient can be computed in time that scales linearly in the size of the PC, allowing fast gradient-based algorithms for learning NNs. Third, our method trains an NN in a self-supervised manner without having to rely on pre-computed solutions to arbitrary MMAP problems, thus circumventing the need to solve intractable MMAP problems in practice. Fourth, we demonstrate via a large-scale experimental evaluation that our proposed NN approximator yields higher quality MMAP solutions as compared to existing approximate schemes.

\section{Preliminaries}
\begin{figure*}
    \centering
    \begin{subfigure}[b]{0.45\textwidth}
    \resizebox{0.95\textwidth}{!}{
    \begin{tikzpicture}[thick,scale=\myscale]
    \node[circle,draw,label=north:({\footnotesize 0.0778})](r) {$+$};
    \node[circle,draw](l11) [below left = of r,label=west:({\footnotesize 0.026})] {$\times$};
    \node[circle,draw](l12) [below right = of r,label=west:({\footnotesize 0.1})] {$\times$};
    \node(l21) [below left = of l11,label=south:({\footnotesize 1})] {$X_1$};
    \node[circle,draw](l22) [below = of l11,label={[xshift=5.5mm]north:({\footnotesize 0.026})}] {$+$};
    \node[circle,draw](l23) [below = of l12,,label={[xshift=-4mm]north:({\footnotesize 0.1})}] {$+$};
    \node(l24) [below right = of l12,label=south:({\footnotesize 1})] {$\neg X_1$};
    \node[circle,draw](l31) [below left = of l22,label=west:({\footnotesize 0.03})] {$\times$};
    \node[circle,draw](l32) [below left = of l23,label=west:({\footnotesize 0.02})] {$\times$};
    \node[circle,draw](l33) [below right = of l23,label=west:({\footnotesize 0.12})] {$\times$};
    \node[circle,draw](l41) [below left = of l31,label=north:({\footnotesize 0.3})] {$+$};
    \node(l42) [below  = of l31,label=south:({\footnotesize 1})] {$\neg X_2$};
    \node[circle,draw](l43) [right = of l42,label=north:({\footnotesize 0.1})] {$+$};
    \node(l44) [right = of l43,label=south:({\footnotesize 1})] {$X_2$};
    \node[circle,draw](l45) [right = of l44,label=north:({\footnotesize 0.2})] {$+$};
    \node(l46) [right = of l45,label=south:({\footnotesize 1})] {$\neg X_2$};
    \node[circle,draw](l47) [right = of l46,label=north:({\footnotesize 0.6})] {$+$};
    \node(l51) [below left = of l41,label=south:({\footnotesize 1})] {$X_3$};
    \node(l52) [right  = of l51,label=south:({\footnotesize 0})] {$\neg X_3$};
    \node(l53) [right = of l52,label=south:({\footnotesize 0})] {$X_4$};
    \node(l54) [right = of l53,label=south:({\footnotesize 1})] {$\neg X_4$};
    \node(l55) [right = of l54,label=south:({\footnotesize 1})] {$X_3$};
    \node(l56) [right = of l55,label=south:({\footnotesize 0})] {$\neg X_3$};
    \node(l57) [right = of l56,label=south:({\footnotesize 0})] {$X_4$};
    \node(l58) [right = of l57,label=south:({\footnotesize 1})] {$\neg X_4$};
    
    \draw[->] (r) edge node[left, xshift=-1mm]{$0.3$} (l11);
    \draw[->] (r) edge node[right, xshift=1mm]{$0.7$}(l12);
    \draw[->] (l11) edge (l21);
    \draw[->] (l11) edge (l22);
    \draw[->] (l12) edge (l23);
    \draw[->] (l12) edge (l24);
    \draw[->] (l22) edge node[left, xshift=-1mm]{$0.6$} (l31);
    \draw[->] (l22) edge node[right, xshift=1mm]{$0.4$} (l32);
    \draw[->] (l23) edge node[left, xshift=-1mm]{$0.2$}(l32);
    \draw[->] (l23) edge node[right, xshift=1mm]{$0.8$} (l33);

    \draw[->] (l31) edge (l41);
    \draw[->] (l31) edge (l42);
    \draw[->] (l31) edge (l43);

    \draw[->] (l32) edge (l43);
    \draw[->] (l32) edge (l44);
    \draw[->] (l32) edge (l45);

    \draw[->] (l33) edge (l45);
    \draw[->] (l33) edge (l46);
    \draw[->] (l33) edge (l47);

    \draw[->] (l41) edge node[left, xshift=-1mm]{$0.3$}(l51);
    \draw[->] (l41) edge node[right, xshift=1mm]{$0.7$}(l52);
    \draw[->] (l43) edge node[left, xshift=-1mm]{$0.9$}(l53);
    \draw[->] (l43) edge node[right, xshift=1mm]{$0.1$}(l54);
    \draw[->] (l45) edge node[left, xshift=-1mm]{$0.2$}(l55);
    \draw[->] (l45) edge node[right, xshift=1mm]{$0.8$}(l56);
    \draw[->] (l47) edge node[left, xshift=-1mm]{$0.4$}(l57);
    \draw[->] (l47) edge node[right, xshift=1mm]{$0.6$}(l58);
    \eat{
    \node(r) at (0,6) {$+$};
    \node(l1) at (1,5){$\times$};
    \node(l1) at (1,7){$\times$};
    }
    \eat{
    \node (a) {$A$};
    \node (z) [right = of a] {$Z$};
    \node (b) [right = of z] {$B$};
    \node (x) [below left = of z] {$X$};
    \node (y) [below right = of z] {$Y$};
    \node (c) [below right = of x] {$C$};
    \path (a) edge (x);
    \path (a) edge (z);
    \path (b) edge (y);
    \path (b) edge (z);
    \path (c) edge (x);
    \path (c) edge (y);
    \path (x) edge (y);
    \path (z) edge (x);
    \path (z) edge (y); 
    }
    \end{tikzpicture}
}
    \caption{PC}
    \end{subfigure}
    \begin{subfigure}[b]{0.45\textwidth}
    \resizebox{0.95\textwidth}{!}{
    \begin{tikzpicture}[thick,scale=\myscale]
    \node[circle,draw,label=north:({\footnotesize 0.0832})](r) {$+$};
    \node[circle,draw](l11) [below left = of r,label=west:({\footnotesize 0.03707})] {$\times$};
    \node[circle,draw](l12) [below right = of r,label=west:({\footnotesize 0.103})] {$\times$};
    \node(l21) [below left = of l11,label=south:({\footnotesize 1})] {$X_1$};
    \node[circle,draw](l22) [below = of l11,label={[xshift=5.5mm]north:({\footnotesize 0.037})}] {$+$};
    \node[circle,draw](l23) [below = of l12, label={[xshift=-5.5mm]north:({\footnotesize 0.103})}] {$+$};

    \node(l24) [below right = of l12,label=south:({\footnotesize 1})] {$\neg X_1$};
    \node[circle,draw](l31) [below left = of l22,label=west:({\footnotesize 0.04256})] {$\times$};
    \node[circle,draw](l32) [below left = of l23,label=west:({\footnotesize 0.02884})] {$\times$};
    \node[circle,draw](l33) [below right = of l23,label=west:({\footnotesize 0.1215})] {$\times$};
    \node[circle,draw](l41) [below left = of l31,label=north:({\footnotesize 0.304})] {$+$};
    \node(l42) [below  = of l31,label=south:({\footnotesize 1})] {$\neg X_2$};
    \node[circle,draw](l43) [right = of l42,label=north:({\footnotesize 0.14})] {$+$};
    \node(l44) [right = of l43,label=south:({\footnotesize 1})] {$X_2$};
    \node[circle,draw](l45) [right = of l44, label={[xshift=-2mm, yshift=0.6mm]north:({\footnotesize 0.206})}] {$+$};

    \node(l46) [right = of l45,label=south:({\footnotesize 1})] {$\neg X_2$};
    \node[circle,draw](l47) [right = of l46,label=north:({\footnotesize 0.59})] {$+$};
    \node(l51) [below left = of l41,label=south:({\footnotesize 0.99})] {$X_3^c$};
    \node(l52) [right  = of l51,label=south:({\footnotesize 0.01})] {$\neg X_3^c$};
    \node(l53) [right = of l52,label=south:({\footnotesize 0.05})] {$X_4^c$};
    \node(l54) [right = of l53,label=south:({\footnotesize 0.95})] {$\neg X_4^c$};
    \node(l55) [right = of l54,label=south:({\footnotesize 0.99})] {$X_3^c$};
    \node(l56) [right = of l55,label=south:({\footnotesize 0.01})] {$\neg X_3^c$};
    \node(l57) [right = of l56,label=south:({\footnotesize 0.05})] {$X_4^c$};
    \node(l58) [right = of l57,label=south:({\footnotesize 0.95})] {$\neg X_4^c$};
    
    \draw[->] (r) edge node[left, xshift=-1mm]{$0.3$} (l11);
    \draw[->] (r) edge node[right, xshift=1mm]{$0.7$}(l12);
    \draw[->] (l11) edge (l21);
    \draw[->] (l11) edge (l22);
    \draw[->] (l12) edge (l23);
    \draw[->] (l12) edge (l24);
    \draw[->] (l22) edge node[left, xshift=-1mm]{$0.6$} (l31);
    \draw[->] (l22) edge node[right, xshift=1mm]{$0.4$} (l32);
    \draw[->] (l23) edge node[left, xshift=-1mm]{$0.2$}(l32);
    \draw[->] (l23) edge node[right, xshift=1mm]{$0.8$} (l33);

    \draw[->] (l31) edge (l41);
    \draw[->] (l31) edge (l42);
    \draw[->] (l31) edge (l43);

    \draw[->] (l32) edge (l43);
    \draw[->] (l32) edge (l44);
    \draw[->] (l32) edge (l45);

    \draw[->] (l33) edge (l45);
    \draw[->] (l33) edge (l46);
    \draw[->] (l33) edge (l47);

    \draw[->] (l41) edge node[left, xshift=-1mm]{$0.3$}(l51);
    \draw[->] (l41) edge node[right, xshift=1mm]{$0.7$}(l52);
    \draw[->] (l43) edge node[left, xshift=-1mm]{$0.9$}(l53);
    \draw[->] (l43) edge node[right, xshift=1mm]{$0.1$}(l54);
    \draw[->] (l45) edge node[left, xshift=-1mm]{$0.2$}(l55);
    \draw[->] (l45) edge node[right, xshift=1mm]{$0.8$}(l56);
    \draw[->] (l47) edge node[left, xshift=-1mm]{$0.4$}(l57);
    \draw[->] (l47) edge node[right, xshift=1mm]{$0.6$}(l58);
    \eat{
    \node(r) at (0,6) {$+$};
    \node(l1) at (1,5){$\times$};
    \node(l1) at (1,7){$\times$};
    }
    \eat{
    \node (a) {$A$};
    \node (z) [right = of a] {$Z$};
    \node (b) [right = of z] {$B$};
    \node (x) [below left = of z] {$X$};
    \node (y) [below right = of z] {$Y$};
    \node (c) [below right = of x] {$C$};
    \path (a) edge (x);
    \path (a) edge (z);
    \path (b) edge (y);
    \path (b) edge (z);
    \path (c) edge (x);
    \path (c) edge (y);
    \path (x) edge (y);
    \path (z) edge (x);
    \path (z) edge (y); 
    }
    \end{tikzpicture}
}
    \caption{QPC}
    \end{subfigure}
    \caption{(a) An example smooth and decomposable PC. The figure also shows value computation for answering the query $\text{p}_\mathcal{M}(X_3=1,X_4=0)$. The values of the leaf, sum, and product nodes are given in parentheses on their bottom, top, and left, respectively. The value of the root node is the answer to the query. (b) QPC obtained from the PC given in (a) for query variables $\{X_3,X_4\}$. For simplicity, here, we use an MMAP problem without any evidence. This is because a given evidence can be incorporated into the
PC by appropriately setting the leaf nodes. We also show value computations for the following leaf initialization: $X_3^c=0.99,\neg X_3^c=0.01, X_4^c=0.05,\neg X_4^c=0.95$ and all other leaves are set to $1$.}
    \label{fig:example}
\end{figure*}
We use upper case letters (e.g., $X$) to denote random variables and corresponding lower case letters (e.g., $x$) to denote an assignment of a value to a variable. We use bold upper case letters (e.g., $\mathbf{X}$) to denote a set of random variables and corresponding bold lower case letters (e.g., $\mathbf{x}$) to denote an assignment of values to all variables in the set. Given an assignment $\mathbf{x}$ to all variables in $\mathbf{X}$ and a variable $Y \in \mathbf{X}$, let $\mathbf{x}_Y$ denote the projection of $\mathbf{x}$ on $Y$. We assume that all random variables take values from the set $\{0, 1\}$; although note that it is easy to extend our method to multi-valued variables. 

\subsection{Probabilistic Circuits} 
A probabilistic circuit (PC) $\mathcal{M}$ \cite{choiProbCirc20} defined over a set of variables $\mathbf{X}$ represents a joint probability distribution over $\mathbf{X}$ using a rooted directed acyclic graph. The graph consists of three types of nodes: internal sum nodes that are labeled by $+$, internal product nodes that are labeled by $\times$, and leaf nodes that are labeled by either $X$ or $\neg X$ where $X \in \mathbf{X}$. Sum nodes represent conditioning, and an edge into a sum node $n$ from its child node $m$ is labeled by a real number $\omega(m,n)>0$. Given an internal node (either a sum or product node) $n$, let $\texttt{ch}(n)$ denote the set of children of $n$. We assume that each sum node $n$ is normalized and satisfies the following property: $\sum_{m\in \texttt{ch}(n)}\omega(m,n) = 1$. 

In this paper, we focus on a class of PCs which are \textit{smooth and decomposable} \cite{choiProbCirc20,NEURIPS2021_vergari}. Examples of such PCs include sum-product networks \cite{poon2011sum,rahman&gogate16b}, mixtures of cutset networks \cite{rahman2014cutset,rahman_cutsetnetworks_2016}, and arithmetic circuits obtained by compiling probabilistic graphical models \cite{darwiche2003differential}. These PCs admit tractable marginal inference, a key property that we leverage in our proposed method.

\begin{definition}
We say that a sum or a product node $n$ is defined over a variable $X$ if there exists a directed path from $n$ to a leaf node labeled either by $X$ or $\neg X$. A PC is \textbf{\textit{smooth}} if each sum node is such that its children are defined over the same set of variables. A PC is \textbf{\textit{decomposable}} if each product node is such that its children are defined over disjoint subsets of variables.
\end{definition}

\begin{example}
Figure \ref{fig:example}(a) shows a smooth and decomposable probabilistic circuit defined over $\mathbf{X}=\{X_1,\ldots,X_4\}$.
\end{example}

\subsection{Marginal Inference in PCs}
Next, we describe how to compute the probability of an assignment to a subset of variables in a smooth and decomposable PC. This task is called the marginal inference (\texttt{MAR}) task. We begin by describing some additional notation.

Given a PC $\mathcal{M}$ defined over $\mathbf{X}$, let $\mathcal{S}$, $\mathcal{P}$ and $\mathcal{L}$ denote the set of sum, product and leaf nodes of $\mathcal{M}$ respectively. Let $\mathbf{Q} \subseteq \mathbf{X}$. Given a node $m$ and an assignment $\mathbf{q}$, let $v(m,\mathbf{q})$ denote the value of $m$ given $\mathbf{q}$. Given a leaf node $n$, let $\texttt{var}(n)$ denote the variable associated with $n$ and let $l(n,\mathbf{q})$ be a function, which we call \textit{leaf function}, that is defined as follows. $l(n,\mathbf{q})$ equals $0$ if any of the following two conditions are satisfied: (1) the label of $n$ is $Q$ where $Q \in \mathbf{Q}$ and $\mathbf{q}$ contains the assignment $Q=0$; and (2) if the label of $n$ is $\neg Q$ and $\mathbf{q}$ contains the assignment $Q=1$. Otherwise, it is equal to $1$. Intuitively, the leaf function assigns all leaf nodes that are inconsistent with the assignment $\mathbf{q}$ to $0$ and the remaining nodes, namely those that are consistent with $\mathbf{q}$ and those that are not part of the query to $1$.

Under this notation, and given a leaf function $l(n,\mathbf{q})$, the marginal probability of any assignment $\mathbf{q}$ w.r.t $\mathcal{M}$, and denoted by $\text{p}_{\mathcal{M}}(\mathbf{q})$ can be computed by performing the following recursive \textit{value computations}:
\begin{align}
\footnotesize
\label{eq:recursion}
    v(n,\mathbf{q}) = \left \{ \begin{array}{ll}
    l(n,\mathbf{q}) & \text{if $n \in \mathcal{L}$}\\
    \sum_{m \in \texttt{ch}(n)} \omega(m,n) v(m,\mathbf{q}) & \text{if $n \in \mathcal{S}$}\\
    \prod_{m \in \texttt{ch}(n)} v(m,\mathbf{q}) &\text{if $n \in \mathcal{P}$}
    \end{array} \right .
\end{align}
Let $r$ denote the root node of $\mathcal{M}$. Then, the probability of $\mathbf{q}$ w.r.t. $\mathcal{M}$, denoted by $\text{p}_\mathcal{M}(\mathbf{q})$ equals $v(r,\mathbf{q})$. Note that if $\mathbf{Q}=\mathbf{X}$, then $v(r,\mathbf{x})$ denotes the probability of the joint assignment $\mathbf{x}$ to all variables in the PC. Thus
\[ \footnotesize v(r,\mathbf{q})= \sum_{\mathbf{y} \in \{0,1\}^{|\mathbf{Y}|}} v (r,(\mathbf{q},\mathbf{y})) \]
where $\mathbf{Y}=\mathbf{X} \setminus \mathbf{Q}$ and the notation $(\mathbf{q},\mathbf{y})$ denotes the \textit{composition} of the assignments to $\mathbf{Q}$ and $\mathbf{Y}$ respectively.

Since the recursive value computations require only one bottom-up pass over the PC, \texttt{MAR} inference is tractable or linear time in smooth and decomposable PCs.

\begin{example}
Figure \ref{fig:example}(a) shows bottom-up, recursive value computations for computing the probability of the assignment $(X_3=1,X_4=0)$ in our running example. Here, the leaf nodes $\neg X_3$ and $X_4$ are assigned to $0$ and all other leaf nodes are assigned to $1$. The number in parentheses at the top, left, and bottom of each sum, product and leaf nodes respectively shows the value of the corresponding node. The value of the root node equals $\text{p}_\mathcal{M}(X_3=1,X_4=0)$.
\end{example}
\subsection{Marginal Maximum-a-Posteriori (MMAP) Inference in PCs}
Given a PC $\mathcal{M}$ defined over $\mathbf{X}$, let $\mathbf{E} \subseteq \mathbf{X}$ and $\mathbf{Q} \subseteq \mathbf{X}$ denote the set of evidence and query variables respectively such that $\mathbf{E} \cap \mathbf{Q}=\emptyset$. Let $\mathbf{H}=\mathbf{X} \setminus (\mathbf{Q} \cup \mathbf{E})$ denote the set of hidden variables. Given an assignment $\mathbf{e}$ to the evidence variables (called evidence), the MMAP task seeks to find an assignment $\mathbf{q}$ to $\mathbf{Q}$ such that the probability of the assignment $(\mathbf{e},\mathbf{q})$ w.r.t. $\mathcal{M}$ is maximized. Mathematically, 
\begin{align}
\footnotesize
\label{eqn:mmap}
\texttt{MMAP}(\mathbf{Q},\mathbf{e}) =& \argmax_{\mathbf{q}}  \text{p}_{\mathcal{M}}(\mathbf{e},\mathbf{q})\\ =& \argmax_{\mathbf{q}}  \sum_{\mathbf{h} \in \{0,1\}^{|\mathbf{H}|}}\text{p}_{\mathcal{M}}(\mathbf{e},\mathbf{q},\mathbf{h})
\end{align}
If $\mathbf{H} =\emptyset$ (namely $\mathbf{Q}$ is the set of non-evidence variables), then MMAP corresponds to the most probable explanation (MPE) task. It is known that both MMAP and MPE tasks are at least NP-hard in smooth and decomposable PCs \cite{park&darwiche04,decamposNewComplexityResultsMAPBayesianNetworks,peharz2015foundations}, and even NP-hard to approximate \cite{conaty17,mei2018maximum}. 

A popular approach to solve the MMAP task in PCs is to replace the sum  ($\sum$) operator with the max operator during bottom-up, recursive value computations and then performing a second top-down pass to find the assignment \cite{poon2011sum}. 

\section{A Neural Optimizer for MMAP in PC\text{s}}\label{sec:method}
In this section, we introduce a learning-based approach using deep neural networks (NNs) to approximately solve the MMAP problem in PCs. Formally, the NN represents a function $f_\theta(.)$ that is parameterized by $\theta$, and takes an assignment $\mathbf{e}$ over the evidence variables as input and outputs an assignment $\mathbf{q}$ over the query variables. Our goal is to design \emph{generalizable, continuous loss functions} for updating the parameters of the NN such that once learned, at test time, given an assignment $\mathbf{e}$ to the evidence variables as input, the NN outputs near-optimal solutions to the MMAP problem.

In this paper, we assume that the sets of evidence $\left (\mathbf{E} = \{E_i\}_{i=1}^N \right )$ and query $\left (\mathbf{Q}=\{Q_j\}_{j=1}^M \right )$ variables are known \textit{a priori} and do not change at both training and test time. We leave as future work the generalization of our approach that can handle variable length, arbitrarily chosen evidence, and query sets. Also, note that our proposed method does not depend on the particular NN architecture used, and we only require that each output node is a continuous quantity in the range $[0,1]$ and uses a differentiable activation function (e.g., the sigmoid function). 

We can learn the parameters of the given NN either in a \textit{supervised} manner or in a \textit{self-supervised} manner. However, the supervised approach is impractical, as described below. 

In the supervised setting, we assume that we are given training data $\mathcal{D} = \{\langle \mathbf{e}_1, \mathbf{q}_1^*\rangle,\ldots,\langle \mathbf{e}_d, \mathbf{q}_d^*\rangle\}$, where each input $\mathbf{e}_i$ is an assignment to the evidence variables, and each \eat{output} (label) $\mathbf{q}_i^*$ is an optimal solution to the corresponding MMAP task, namely $\mathbf{q}_i^*=\texttt{MMAP}(\mathbf{Q},\mathbf{e}_i)$. We then use supervised loss functions such as the mean-squared-error (MSE) $\sum_{i=1}^d\|\mathbf{q}^*_i - \mathbf{q}_i^c)\|_2^2/d$ and the mean-absolute-error (MAE) $\sum_{i=1}^d\|\mathbf{q}^*_i - \mathbf{q}_i^c\|_1/d$ where $\mathbf{q}_i^c$ is the predicted assignment (note that $\mathbf{q}_i^c$ is continuous), and standard gradient-based methods to learn the parameters. Although supervised approaches allow us to use simple-to-implement loss functions, they are \textit{impractical} if the number of query variables is large because they require access to the exact solutions to several intractable MMAP problems\footnote{Note that the training data used to train the NN in the supervised setting is different from the training data used to learn the PC. In particular, in the data used to train the PC, the assignments to the query variables $\mathbf{Q}$ may not be optimal solutions of $\texttt{MMAP}(\mathbf{Q},\mathbf{e})$.}. We therefore propose to use a \textit{self-supervised approach}. 

\subsection{A Self-Supervised Loss Function for PCs}
\label{sec:ssl_loss}
In the self-supervised setting, we need access to training data in the form of assignments to the evidence variables, i.e., $\mathcal{D}' = \{\mathbf{e}_1,\ldots,\mathbf{e}_d\}$. Since smooth and decomposable PCs admit perfect sampling, these assignments can be easily sampled from the PC via top-down AND/OR sampling \cite{gogate&dechter12}. The latter yields an assignment $\mathbf{x}$ over all the random variables in the PC. Then we simply project $\mathbf{x}$ on the evidence variables $\mathbf{E}$ to yield a training example $\mathbf{e}$. Because each training example can be generated in time that scales linearly with the size of the PC, in practice, our proposed self-supervised approach is likely to have access to much larger number of training examples compared to the supervised approach.

Let $\mathbf{q}^c$ denote the MMAP assignment predicted by the NN given evidence $\mathbf{e} \in \mathcal{D}'$ where $\mathbf{q}^c \in [0,1]^{M}$. In MMAP inference, given $\mathbf{e}$, we want to find an assignment $\mathbf{q}$ such that $\ln \texttt{p}_{\mathcal{M}}(\mathbf{e},\mathbf{q})$ is maximized, namely, $-\ln \texttt{p}_{\mathcal{M}}(\mathbf{e},\mathbf{q})$ is minimized. Thus, a natural loss function that we can use is  $-\ln \texttt{p}_{\mathcal{M}}(\mathbf{e},\mathbf{q})$. Unfortunately, the NN outputs a continuous vector $\mathbf{q}^c$ and as a result $\texttt{p}_{\mathcal{M}}(\mathbf{e},\mathbf{q}^c)$ is not defined. Therefore, we cannot use $-\ln \texttt{p}_{\mathcal{M}}(\mathbf{e},\mathbf{q}^c)$ as a loss function.

One approach to circumvent this issue is to use a threshold (say $0.5$) to convert each continuous quantity in the range [0,1] to a binary one. A problem with this approach is that the threshold function is not differentiable. 

Therefore, we propose to construct a smooth, differentiable loss function that given $\mathbf{q}^c=(q_1^c,\ldots,q_M^c)$ approximates $-\ln \texttt{p}_{\mathcal{M}}(\mathbf{e},\mathbf{q})$ where $\mathbf{q}=(q_1=[q_1^c>0.5],\ldots,q_M=[q_M^c>0.5])$ and $[q_i^c>0.5]$ is an indicator function which is $1$ if $q_i^c>0.5$ and $0$ otherwise. The key idea in our approach is to construct a new PC, which we call \textit{Query-specific PC} (QPC) by replacing all binary leaf nodes associated with the query variables in the original PC, namely those labeled by $Q$ and $\neg Q$ where $Q \in \mathbf{Q}$, with continuous nodes $Q^c \in [0,1]$ and $\neg Q^c \in [0,1]$. Then our proposed loss function is obtained using value computations (at the \textit{root node} of the QPC) via a simple modification of the \textit{leaf function} of the PC. At a high level, our new leaf function assigns each leaf node labeled by $Q_j^c$ such that $Q_j \in \mathbf{Q}$ to its corresponding estimate $q_j^c$, obtained from the NN and each leaf node labeled by $\neg Q_j^c$ such that $Q_j \in \mathbf{Q}$ to $1-q_j^c$. 

Formally, for the QPC, we propose to use leaf function $l'(n,(\mathbf{e},\mathbf{q}^c))$ defined as follows:
\begin{enumerate}
    \item If the label of $n$ is $Q_j^c$ such that $Q_j \in \mathbf{Q}$ then $l'(n,(\mathbf{e},\mathbf{q}^c))=q^c_{j}$.
    \item If $n$ is labeled by $\neg Q_j^c$ such that $Q_j \in \mathbf{Q}$ then $l'(n,(\mathbf{e},\mathbf{q}^c))=1-q^c_{j}$.
    \item If $n$ is labeled by $E_k$ such that $E_k \in \mathbf{E}$ and the assignment $E_k=0$ is in $\mathbf{e}$ then $l'(n,(\mathbf{e},\mathbf{q}^c))=0$.
    \item If $n$ is labeled by $\neg E_k$ such that $E_k \in \mathbf{E}$ and the assignment $E_k=1$ is in $\mathbf{e}$ then $l'(n,(\mathbf{e},\mathbf{q}^c))=0$.
    \item If conditions (1)-(4) are not met then $l'(n,(\mathbf{e},\mathbf{q}^c))=1$.
\end{enumerate}
The value of each node $n$ in the QPC, denoted by $v'(n,(\mathbf{e},\mathbf{q}^c))$ is given by a similar recursion to the one given in Eq. \eqref{eq:recursion} for PCs, except that the leaf function $l(n,\mathbf{q})$ is replaced by the new (continuous) leaf function $l'(n,(\mathbf{e},\mathbf{q}^c))$. Formally, $v'(n,(\mathbf{e},\mathbf{q}^c))$ is given by
\begin{align}
v'(n,(\mathbf{e},\mathbf{q}^c)) = 
\begin{cases} 
l' (n,(\mathbf{e},\mathbf{q}^c)) & \text{if $n \in \mathcal{L}$} \\
\sum_{m \in \texttt{ch}(n)} \omega(m,n) v'(m,(\mathbf{e},\mathbf{q}^c)) & \text{if $n \in \mathcal{S}$} \\
\prod_{m \in \texttt{ch}(n)} v'(n,(\mathbf{e},\mathbf{q}^c)) & \text{if $n \in \mathcal{P}$}
\end{cases}
\label{eq:recursion2}
\end{align}

Let $r$ denote the root node of $\mathcal{M}$, then we propose to use $-\ln v'(r,(\mathbf{e},\mathbf{q}^c))$ as a loss function. 
\begin{example}
    Figure \ref{fig:example}(b) shows the QPC corresponding to the PC shown in Figure \ref{fig:example}(a). We also show value computations for the assignment $(X_3^c=0.99,X_4^c=0.05)$.
\end{example}

\subsection{Tractable Gradient Computation}
Our proposed loss function is smooth and continuous because by construction, it is a negative logarithm of a \textit{multilinear function} over $\mathbf{q}^c$. Next, we show that the partial derivative of the function w.r.t. $q_j^c$ can be computed in linear time in the size of the QPC\footnote{Recall that $q_j^c$ is an output node of the NN and therefore backpropagation over the NN can be performed in time that scales linearly with the size of the NN and the QPC}. More specifically, in order to compute the partial derivative of QPC with respect to $\mathbf{q}_j^c$, we simply have to use a new leaf function which is identical to $l'$ except that if the label of a leaf node $n$ is $Q_j^c$ then we set its value to $1$ (instead of $q_j^c$) and if it is $\neg Q_j^c$ then we set its value $-1$ (instead of $1-q_j^c$). We then perform bottom-up recursive value computations over the QPC and the value of the root node is the partial derivative of the QPC with respect to $\mathbf{q}_j^c$. In summary, it is straight-forward to show that:
\begin{proposition}
    The gradient of the loss function $-\ln v'(r,(\mathbf{e},\mathbf{q}^c))$ w.r.t. $\mathbf{q}_j^c$ can be computed in time and space that scales linearly with the size of $\mathcal{M}$.
\end{proposition}

\begin{example}
The partial derivative of the QPC given in figure \ref{fig:example}(b) w.r.t. $x_3^c$ given $(X_3^c=0.99,X_4^c=0.05)$ can be obtained by setting the leaf nodes $X_3^c$ to $1$ and $\neg X_3^c$ to $-1$, assigning all other leaves to the values shown in Figure \ref{fig:example}(b) and then performing value computations. After the value computation phase, the value of the root node will equal the partial derivative of the QPC w.r.t. $x_3^c$.
\end{example}

\subsection{Improving the Loss Function}
As mentioned earlier, our proposed loss function is a continuous approximation of the discrete function $-\ln v(r,(\mathbf{e},\mathbf{q}))$ where $\mathbf{q}=(q_1=[q_1^c>0.5],\ldots,q_M=[q_M^c>0.5])$ and the difference between the two is minimized iff $\mathbf{q}=\mathbf{q}^c$. Moreover, since the set of continuous assignments includes the discrete assignments, it follows that:
\begin{align} 
\footnotesize
\label{eq:loss-function} 
\min_{\mathbf{q}^c} \left \{ -\ln v'(r,(\mathbf{e},\mathbf{q}^c)) \right \} \leq \min_{\mathbf{q}} \left \{-\ln v(r,(\mathbf{e},\mathbf{q})) \right \}   
\end{align}
Since the right-hand side of the inequality given in \eqref{eq:loss-function} solves the MMAP task, we can improve our loss function by tightening the lower bound. This can be accomplished using an entropy-based penalty, controlled by a hyper-parameter $\alpha>0$, yielding the loss function
\begin{align}
\footnotesize
\nonumber
    \ell(\mathbf{q}^c) =
    -\ln v'(r,(\mathbf{e},\mathbf{q}^c))-\\ 
    \label{eqn:loss} \alpha \sum_{j=1}^{M} 
     q^c_j \log(q^c_j) +  (1-q^c_j) \log(1-q^c_j) 
\end{align}  
The second term in the expression given above is minimized when each $q^c_j$ is closer to $0$ or $1$ and is maximized when $q^c_j=0.5$. Therefore, it encourages 0/1 (discrete) solutions. The hyperparameter $\alpha$ controls the magnitude of the penalty. When $\alpha=0$, the above expression finds an assignment based on the continuous approximation $-\ln v'(r,(\mathbf{e},\mathbf{q}^c))$. On the other hand, when $\alpha=\infty$ then only discrete solutions are possible yielding a non-smooth loss function. $\alpha$ thus helps us trade the smoothness of our proposed loss function with its distance to the true loss.

\section{Experiments}
In this section, we describe and analyze the results of our comprehensive experimental evaluation for assessing the performance of our novel \underline{S}elf-\underline{S}upervised learning based \underline{M}MAP solver for \underline{P}Cs, referred to as \our hereafter. We begin by describing our experimental setup including competing methods, evaluation criteria, as well as NN architectures, datasets, and PCs used in our study.

\subsection{Competing Methods}
\label{sec:exp-methods}
We use \textit{three polytime baseline methods} from the PC and probabilistic graphical models literature \cite{park&darwiche04,poon2011sum}. We also compared the impact of using the solutions computed by the three baseline schemes as well our method \our as initial state for stochastic hill climbing search.

\noindent\textbf{Baseline 1: MAX Approximation (\spn)}. In this scheme \cite{poon2011sum}, the MMAP assignment is derived by substituting sum nodes with max nodes. During the upward pass, a max node produces the maximum weighted value from its children instead of their weighted sum. Subsequently, the downward pass begins from the root and iteratively selects the highest-valued child of a max node (or one of them), along with all children of a product node. 

\noindent\textbf{Baseline 2: Maximum Likelihood Approximation (\mle)} \cite{park&darwiche04}
For each variable $Q \in \mathbf{Q}$, we first compute the marginal distribution $\text{p}_\mathcal{M}(Q|\mathbf{e})$ and then set $Q$ to $\argmax_{j \in \{0,1\}}\text{p}_\mathcal{M}(Q=j|\mathbf{e})$.

\noindent\textbf{Baseline 3: Sequential Approximation (\seq)}
In this scheme \cite{park&darwiche04}, we assign the query variables one by one until no query variables remain unassigned. At each step, we choose an unassigned query variable $Q_j \in \mathbf{Q}$ that maximizes the probability $\text{p}_\mathcal{M}(q_j|\mathbf{e},\mathbf{y})$ for one of its values $q_j$ and assign it to $q_j$ where $\mathbf{y}$ represents the assignment to the previously considered query variables.

\noindent\textbf{Stochastic Hill Climbing Search.} We used the three baselines and our \our method as the initial state in stochastic hill climbing search for MMAP inference described in \cite{park&darwiche04}. The primary goal of this experiment is to assess whether our scheme can assist local search-based \textit{anytime methods} in reaching better solutions than other heuristic methods for initialization. In our experiments, we ran stochastic hill climbing for 100 iterations for each MMAP problem.

\subsection{Evaluation Criteria}
We evaluated the performance of the competing schemes along two dimensions: log-likelihood scores and inference times. Given evidence $\mathbf{e}$ and query answer $\mathbf{q}$, the log-likelihood score is given by $\ln \text{p}_\mathcal{M}(\mathbf{e},\mathbf{q})$.

\subsection{Datasets and Probabilistic Circuits}
We use twenty-two widely used binary datasets from the tractable probabilistic models' literature \cite{lowdLearningMarkovNetwork2010b,haarenMarkovNetworkStructureLearningRandomizedFeature2012,larochelleNeuralAutoregressiveDistributionEstimator,bekkerTractableLearningComplexProbabilityQueries} (we call them TPM datasets) as well as the binarized MNIST \cite{salakhutdinov2008quantitative}, EMNIST \cite{cohenEMNISTExtensionMNISThandwrittenletters2017} and CIFAR-10 \cite{krizhevskyLearningMultipleLayersFeaturesTinyImages} datasets. We used the DeeProb-kit library \cite{loconte2022deeprob} to learn a sum-product network (our choice of PC) for each dataset. The number of nodes in these learned PCs ranges from 46 to 22027.

For each PC and each test example in the 22 TPM datasets, we generated two types of MMAP instances: MPE instances in which $\mathbf{H}$ is empty and MMAP instances in which $\mathbf{H}$ is not empty. We define query ratio, denoted by $qr$, as the fraction of variables that are part of the query set. For MPE, we selected $qr$ from $\{0.1, 0.3, 0.5, 0.6, 0.7, 0.8, 0.9\}$, and for MMAP, we replaced $0.9$ with $0.4$ to avoid small $\mathbf{H}$ and $\mathbf{E}$. For generating MMAP instances, we used 50\% of the remaining variables as evidence variables (and for MPE instances all remaining variables are evidence variables). 

For the MNIST, EMNIST, and CIFAR-10 datasets, we used $qr=0.7$ and generated MPE instances only. More specifically, we used the top 30\% portion of the image as evidence, leaving the bottom 70\% portion as query variables. Also, in order to reduce the training time for PCs, note that for these datasets, we learned a PC for each class, yielding a total of ten PCs for each dataset.

\subsection{Neural Network Optimizers}
For each PC and query ratio combination, we trained a corresponding neural network (NN) using the loss function described in the previous section. Because we have 22 TPM datasets and 7 query ratios for them, we trained 154 NNs for the MPE task and 154 for the MMAP task. For the CIFAR-10, MNIST and EMNIST datasets, we trained 10 NNs, one for each PC (recall that we learned a PC for each class).

Because our learning method does not depend on the specific choice of neural network architectures, we use a fixed neural network architecture across all experiments: fully connected with four
hidden layers having 128, 256, 512, and 1024 nodes respectively. We used ReLU activation in the hidden layers, sigmoid in the output layer, dropout for regularization \cite{srivastava_waypreventneural_2014} and Adam optimizer \cite{kingma_stochasticoptimization_2017} with a standard learning rate scheduler for 50 epochs. All NNs were trained using PyTorch \cite{paszke2019pytorch} on a single NVIDIA A40 GPU. We select a value for the hyperparameter $\alpha$  used in our loss function (see equation \eqref{eqn:loss}) via 5-fold cross validation. 

\subsection{Results on the TPM Datasets}
\begin{table}[hbt!]
\centering
\resizebox{0.5\columnwidth}{!}{%
\begin{tabular}{|c|cccc|cccc|}
\hline
\large
    & \multicolumn{4}{c|}{MPE}                     & \multicolumn{4}{c|}{MMAP}                    \\ \cline{2-9} 
    & \spn & \cellcolor[HTML]{D4D4D4}\ourfig & \mle & \seq & \spn & \cellcolor[HTML]{D4D4D4}\ourfig & \mle & \seq \\ \hline
\spn & 0   & \cellcolor[HTML]{D4D4D4}64 & 33  & 14  & 0   & \cellcolor[HTML]{D4D4D4}46 & 23  & 10  \\
\rowcolor[HTML]{D4D4D4} 
\ourfig  & 88  & 0                          & 96  & 77  & 97  & 0                          & 102 & 82  \\
\mle & 6   & \cellcolor[HTML]{D4D4D4}49 & 0   & 15  & 3   & \cellcolor[HTML]{D4D4D4}34 & 0   & 10  \\
\seq & 105 & \cellcolor[HTML]{D4D4D4}63 & 105 & 0   & 117 & \cellcolor[HTML]{D4D4D4}53 & 117 & 0   \\ \hline
\end{tabular}%
}
\caption{Contingency tables over the competing methods for MPE and MMAP problems. Highlighted values represent results for \our.}
\label{tab:contingency-base}
\end{table}
We summarize our results for the competing schemes (3 baselines and \our) on the 22 TPM datasets using the two contingency tables given in Table \ref{tab:contingency-base}, one for MPE and one for MMAP. Detailed results are provided in the supplementary material. Recall that we generated 154 test datasets each for MPE and MMAP (22 PCs $\times$ 7 $qr$ values). In all contingency tables, the number in the cell $(i,j)$ equals the number of times (out of 154) that the scheme in the $i$-th row was better in terms of average log-likelihood score than the scheme in the $j$-th column. The difference between 154 and the sum of the numbers in the cells $(i,j)$ and $(j,i)$ equals the number of times the scheme in the $i$-th row and $j$-th column had identical log-likelihood scores.

From the MPE contingency table given in Table \ref{tab:contingency-base}, we observe that \our is superior to \spn, \mle, and \seq approximations. The \seq approximation is slightly better than the \spn approximation, and \mle is the worst-performing scheme. For the harder MMAP task, we see a similar ordering among the competing schemes (see Table \ref{tab:contingency-base}) with \our dominating other schemes. In particular, \our outperforms the \spn and \mle approximations in almost two-thirds of the cases and the \seq method in more than half of the cases.     

\begin{table}[hbt!]
\centering
\resizebox{0.5\columnwidth}{!}{%
\begin{tabular}{|c|cccc|cccc|}
\hline
                   & \multicolumn{4}{c|}{MPE}                                 & \multicolumn{4}{c|}{MMAP}                                \\ \cline{2-9} 
\multirow{-2}{*}{} & \spn & \cellcolor[HTML]{D4D4D4}\ourfig & \mle & \seq & \spn & \cellcolor[HTML]{D4D4D4}\ourfig & \mle & \seq \\ \cline{2-9} 
\spn             & 0      & \cellcolor[HTML]{D4D4D4}40    & 13     & 9      & 0      & \cellcolor[HTML]{D4D4D4}27    & 10     & 16     \\
\rowcolor[HTML]{D4D4D4} 
\ourfig              & 93     & 0                             & 99     & 87     & 98     & 0                             & 100    & 86     \\
\mle             & 19     & \cellcolor[HTML]{D4D4D4}37    & 0      & 14     & 12     & \cellcolor[HTML]{D4D4D4}26    & 0      & 17     \\
\seq             & 85     & \cellcolor[HTML]{D4D4D4}44    & 82     & 0      & 89     & \cellcolor[HTML]{D4D4D4}39    & 90     & 0      \\ \hline
\end{tabular}%
}
\caption{Contingency tables for Hill Climbing Search initialized using the competing methods for MPE and MMAP problems. Highlighted values represent results for \our.}
\label{tab:contingency-hill}
\end{table}

We also investigate the effectiveness of \our  and other baseline approaches when employed as initialization strategies for Hill Climbing Search. These findings are illustrated in the contingency tables given in Table \ref{tab:contingency-hill}. The results for \mpe are presented on the left side, while those for \mmap are presented on the right. Notably, \our outperforms all other baseline approaches in nearly two-thirds of the experiments for both \mpe and \mmap tasks. These results demonstrate that \our can serve as an effective initialization technique for anytime local search-based algorithms.

\begin{figure}[thb!]
    \centering
    \begin{subfigure}{0.49\linewidth}
        \centering
        \includegraphics[width=\linewidth]{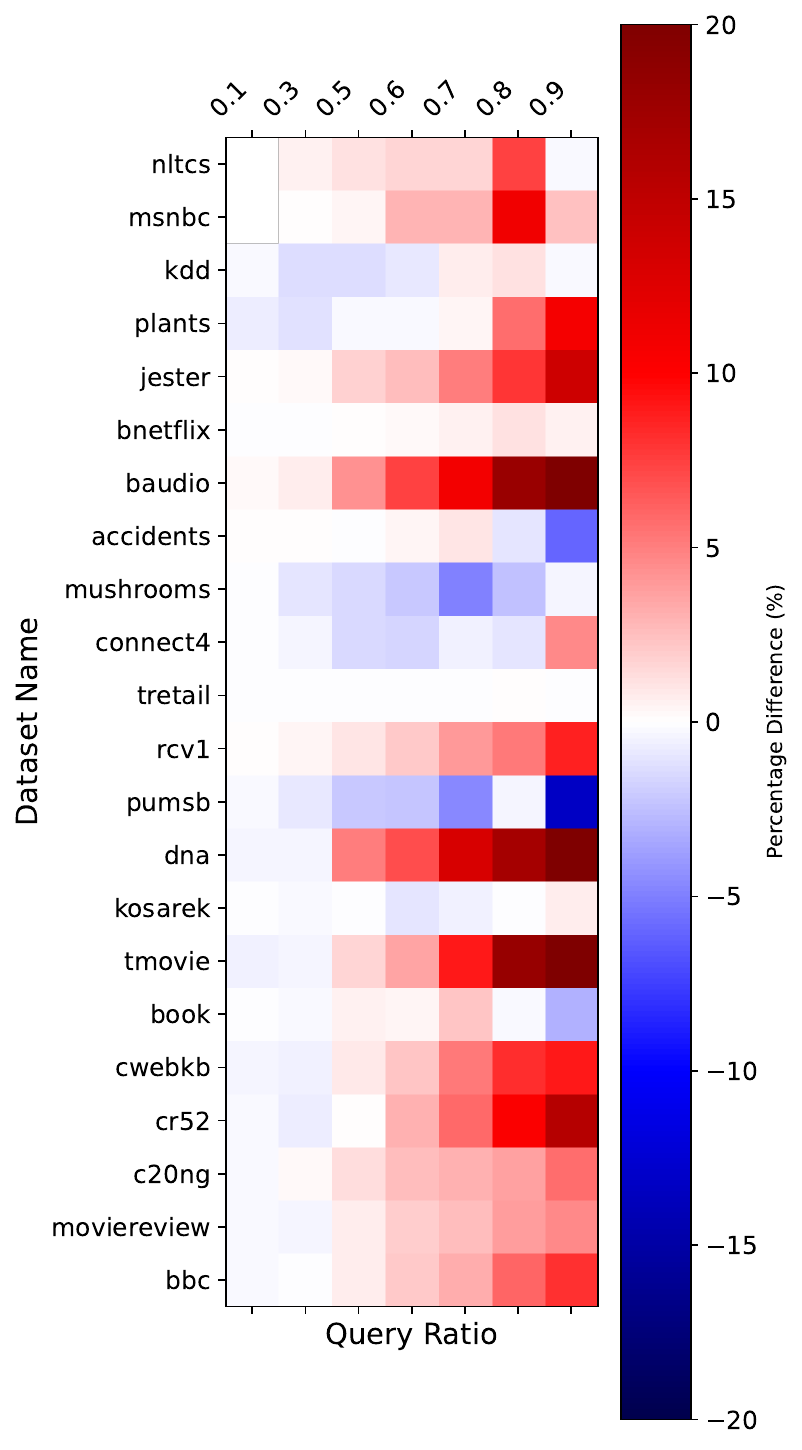}
        \caption{\mpe}
        \label{fig:mpe_nn_vs_spn}
    \end{subfigure}
    \hfill
    \begin{subfigure}{0.49\linewidth}
        \centering
        \includegraphics[width=\linewidth]{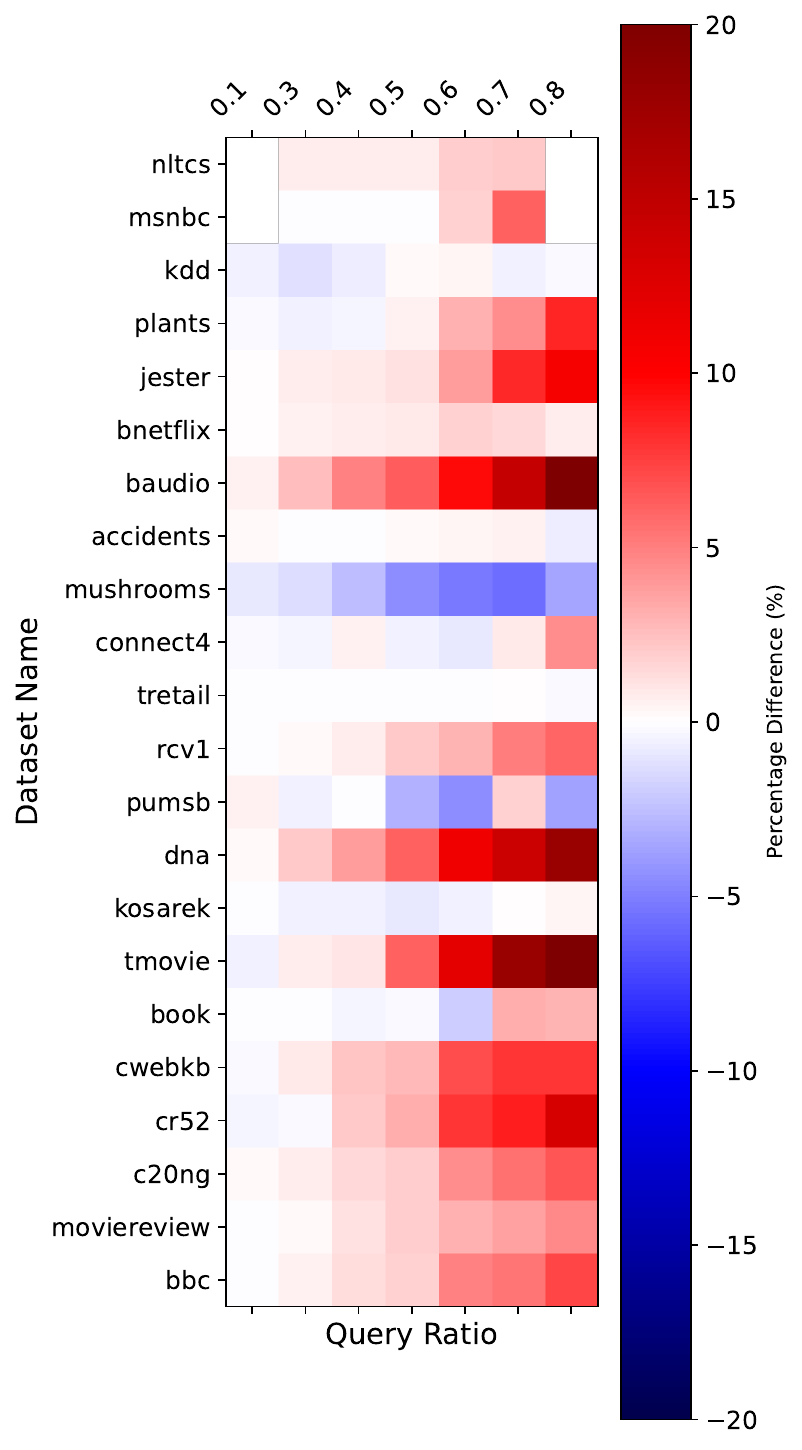}
        \caption{\mmap}
        \label{fig:mmap_nn_vs_spn}
    \end{subfigure}
    
    \caption{Heat map showing the \% difference in log-likelihood scores between \texttt{\our} and \spn approximation. Each row denotes a distinct dataset, with the color gradient depicting the \% Difference. The gradient extends from dark blue to light blue, indicating areas where \spn is superior (negative values), and from light red to dark red, highlighting regions where \texttt{\our} outperforms (positive values).}
    \label{fig:performance_vs_spn}
\end{figure}

In Figure \ref{fig:performance_vs_spn}, via a heat-map representation, we show a more detailed performance comparison between \our and the \spn approximation, which is a widely used baseline for MPE and MMAP inference in PCs. In the heat-map representation, the y-axis represents the datasets (ordered by the number of variables), while the x-axis shows the query ratio. The values in each cell represent the percentage difference between the mean log-likelihood scores of \our and the \spn approximation. Formally, let $ll_{ssmp}$ and $ll_{max}$ denote the mean LL scores of \our and \spn approximation respectively, then the percentage difference is given by
\begin{equation}
\label{eq:percent_diff}
\begin{aligned}
\% \text{Diff.} = \frac{ll_{ssmp}-ll_{max}}{|ll_{max}|} \times 100
\end{aligned}
\end{equation}

From the heatmap for MPE given in Figure \ref{fig:performance_vs_spn}(a), we observe that \our is competitive with the \spn approximation when the size of the query set is small. However, as the number of query variables increases, signaling a more challenging problem, \our consistently outperforms or has similar performance to the \spn method across all datasets, except for accidents, pumsb-star, and book.

The heatmaps for MMAP are illustrated in Figure \ref{fig:performance_vs_spn}(b). We see a similar trend as the one for MPE; \our remains competitive with the \spn approximation, particularly when the number of query variables is small. While \our outperforms (with some exceptions) the \spn approximation when the number of query variables is large.

Finally, we present inference times in the supplement. On average \our requires in the order of 7-10 micro-seconds for MMAP inference on an A40 GPU. The \spn approximation takes 7 milli-seconds (namely, \our is almost 1000 times faster). In comparison, as expected, the \seq and \mle approximations are quite slow, requiring roughly 400 to 600 milliseconds to answer MPE and MMAP queries. In the case of our proposed method (\our), during the inference process, the size of the SPN holds no relevance; its time complexity is linear in the size of the neural network. On the contrary, for the alternative methods, the inference time is intricately dependent on the size of the SPN.

\subsection{Results on the CIFAR-10 Dataset}
\begin{table}[t]
\centering
\resizebox{0.3\columnwidth}{!}{%
\begin{tabular}{|cc
>{\columncolor[HTML]{D4D4D4}}c cc|}
\hline
                                & \spn                       & \ourfig & \mle                       & \seq                       \\ \hline
\spn                             & 0                         & 0       & 0                         & 2                         \\
\cellcolor[HTML]{D4D4D4}\ourfig & \cellcolor[HTML]{D4D4D4}9 & 0       & \cellcolor[HTML]{D4D4D4}9 & \cellcolor[HTML]{D4D4D4}9 \\
\mle                             & 0                         & 0       & 0                         & 2                         \\
\seq                             & 7                         & 0       & 7                         & 0                         \\ \hline
\end{tabular}%
}
\caption{Contingency table over the competing methods for MPE on CIFAR-10. Highlighted values are for \our.}
\label{tab:cifar10-base}
\end{table}
We binarized the CIFAR-10 dataset \eat{\cite{krizhevskyLearningMultipleLayersFeaturesTinyImages}} using a variational autoencoder having 512 bits. We then learned a PC for each of the 10 classes; namely, we learned a PC conditioned on the class variable. As mentioned earlier, we randomly set $70\%$ of the variables as query variables. The contingency table for CIFAR-10 is shown in Table \ref{tab:cifar10-base}. We observe that \our dominates all competing methods while the \seq approximation is the  second-best performing scheme (although note that \seq is computationally expensive).

\subsection{Results on the MNIST and EMNIST Datasets}
Finally, we evaluated \our on the image completion task using the Binarized MNIST \cite{salakhutdinov2008quantitative} and the EMNIST datasets \cite{cohenEMNISTExtensionMNISThandwrittenletters2017}. As mentioned earlier, we used the top 30\% of the image as evidence and estimated the bottom 70\% by solving the MPE task over PCs using various competing methods. 
\begin{table}[t]
\centering
\resizebox{0.5\columnwidth}{!}{%
\begin{tabular}{|c|cccc|cccc|}
\hline
        & \multicolumn{4}{c|}{MNIST}                                                                                                         & \multicolumn{4}{c|}{EMNIST}                                                                                                        \\ \cline{2-9} 
        & \multicolumn{1}{c}{\spn} & \multicolumn{1}{c}{\cellcolor[HTML]{D4D4D4}\ourfig} & \multicolumn{1}{c}{\mle} & \multicolumn{1}{c|}{\seq} & \multicolumn{1}{c}{\spn} & \multicolumn{1}{c}{\cellcolor[HTML]{D4D4D4}\ourfig} & \multicolumn{1}{c}{\mle} & \multicolumn{1}{c|}{\seq} \\ \hline
\spn     & 0                       & \cellcolor[HTML]{D4D4D4}1                           & 0                       & 1                        & 0                       & \cellcolor[HTML]{D4D4D4}1                           & 0                       & 5                        \\
\rowcolor[HTML]{D4D4D4} 
\ourfig & 9                       & 0                                                   & 9                       & 9                        & 7                       & 0                                                   & 7                       & 7                        \\
\mle     & 0                       & \cellcolor[HTML]{D4D4D4}1                           & 0                       & 1                        & 0                       & \cellcolor[HTML]{D4D4D4}1                           & 0                       & 5                        \\
\seq     & 9                       & \cellcolor[HTML]{D4D4D4}1                           & 9                       & 0                        & 3                       & \cellcolor[HTML]{D4D4D4}1                           & 3                       & 0                        \\ \hline
\end{tabular}%
}
\caption{Contingency tables over the competing methods for MPE on the MNIST and EMNIST datasets. Highlighted values represent results for \our.}
\label{tab:mnist-base}
\end{table}
The contingency tables for the MNIST and EMNIST datasets are shown in Table \ref{tab:mnist-base}. We observe that on the MNIST dataset, \our is better than all competing schemes on 9 out of the 10 PCs, while it is inferior to all on one of them. On the EMNIST dataset, \our is better than all competing schemes on 7 out of the 10 PCs and inferior to all on one of the PCs. Detailed results on the image datasets, including qualitative comparisons, are provided in the supplement.

In summary, we find that, on average, our proposed method (\our) is better than other baseline MPE/MMAP approximations in terms of log-likelihood score. Moreover, it is substantially better than the baseline methods when the number of query variables is large. Also, once learned from data, it is also significantly faster than competing schemes.

\section{Conclusion and Future Work}
\eat{In this paper we presented a machine learning based method to solve the MMAP task over PCs. 
\textbf{Neural Approximators for MPE over intractable models}}
In this paper, we introduced a novel self-supervised learning algorithm for solving MMAP queries in PCs. Our contributions comprise a neural network approximator and a self-supervised loss function which leverages the tractability of PCs for achieving scalability. Notably, our method employs minimal hyperparameters, requiring only one in the discrete case. We conducted a comprehensive empirical evaluation across various benchmarks; specifically, we experimented with 22 binary datasets used in tractable probabilistic models community and three classic image datasets, MNIST, EMNIST, and CIFAR-10. We compared our proposed neural approximator to polytime baseline techniques and observed that it is superior to the baseline methods in terms of log-likelihood scores and is significantly better in terms of computational efficiency. Additionally, we evaluated how our approach performs when used as an initialization scheme in stochastic hill climbing (local) search and found that it improves the quality of solutions output by anytime local search schemes. Our empirical results clearly demonstrated the efficacy of our approach in both accuracy and speed. 

Future work includes compiling PCs to neural networks for answering more complex queries that involve constrained optimization; developing sophisticated self-supervised loss functions; learning better NN architecture for the given PC; generalizing our approach to arbitrarily chosen query and evidence subsets; etc.

\section*{Acknowledgements}
This work was supported in part by the DARPA Perceptually-Enabled Task Guidance (PTG) Program under contract number HR00112220005, by the DARPA Assured Neuro Symbolic Learning and Reasoning (ANSR) under contract number HR001122S0039 and by the National Science Foundation grant IIS-1652835.

\bibliographystyle{unsrtnat}
\bibliography{ref}

\newpage
\appendix
\onecolumn
\section{Proof Sketch for Proposition 1}
Without loss of generality, assume that the root node of the QPC is a sum node, the QPC has alternating levels of sum and product nodes and assume that we are computing the partial derivative of the QPC w.r.t. node $X_j^c$ such that $X_j \in \mathbf{Q}_j$. Because the QPC is smooth, all child nodes of the root node must contain $X_j^c.$ Therefore, by sum rule of derivative, the derivative at the root node is given by:
\begin{align}
    \label{pf:eq1}
\frac{\partial v(r,\mathbf{e},\mathbf{q})}{\partial x_j^c} = \sum_{m \in \texttt{ch}(r)} \omega(m,r) \frac{\partial v(m,\mathbf{e},\mathbf{q})}{\partial x_j^c}
\end{align} 
We assume that the QPC is decomposable. As a result, at each product node $n$ of the root, exactly one child node of $n$ will be defined over $X_j^c$. Therefore, while computing the derivative, we can treat the values of other child nodes as constants (since they do not depend on $X_j^c$). Let the child node of $n$ that is defind over $X_j^c$ be denoted by $m_j$. Then, the partial derivative of the product node w.r.t. $X_j^c$ is given by:
\begin{align}
\label{pf:eq2} \frac{\partial v(n,\mathbf{e},\mathbf{q})}{\partial x_j^c} = \frac{\partial v(m_j,\mathbf{e},\mathbf{q})}{\partial x_j^c}\prod_{m \in \texttt{ch}(n): m  \neq m_j} v(m,\mathbf{e},\mathbf{q}) 
\end{align}

Continuing this analysis further, it is easy to see that the partial derivative of each sum and product node that mentions $X_j^c$ (w.r.t. $X_j^c$) is given by equations \eqref{pf:eq1} and \eqref{pf:eq2} respectively. 

The derivative of the leaf node labeled by $X_j^c$ w.r.t. $x_j^c$ equals $1$, because it is assigned to $x_j^c$. Similarly, the  derivative of the leaf node labeled by $\neg X_j^c$ w.r.t. $x_j^c$ equals $-1$, because it is assigned to $1-x_j^c$.

Thus, we observe that the only leaf assignments that change from the QPC used to compute the loss function are the ones labeled by  $X_j^c$ and $\neg X_j^c$. In particular, they go from $x_j^c$ and $1-x_j^c$ to $1$ and $-1$ respectively. (While all other leaf assignments stay the same.) 

Therefore, the partial derivative of the QPC w.r.t. $X_j^c$ can be computed via value computations using the new leaf assignments described above. Since the value computations require only one pass over the QPC, they run in linear time in the size of the QPC.
\qed

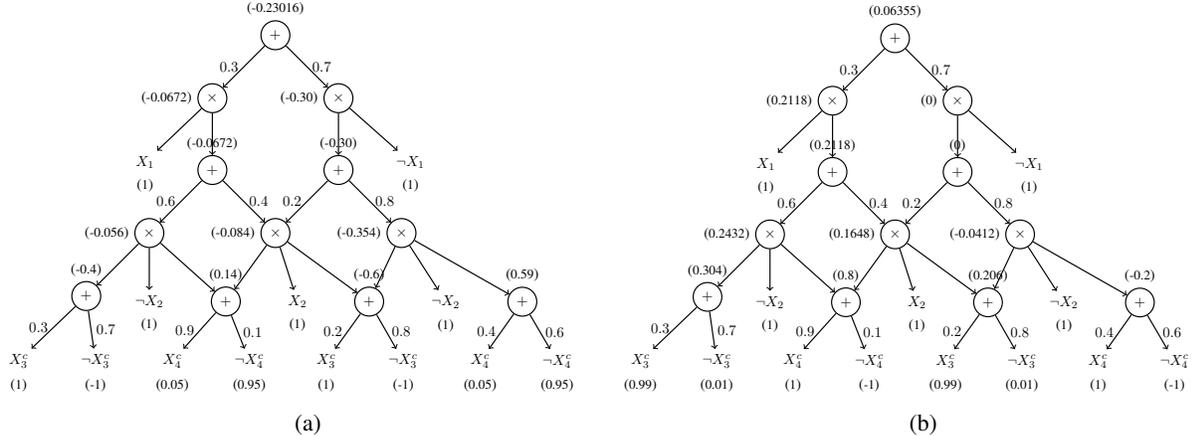
\begin{figure*}[ht!]
    \centering
    \begin{subfigure}[b]{0.49\textwidth}
    \resizebox{0.95\textwidth}{!}{
    \begin{tikzpicture}[thick,scale=\myscale]
    \node[circle,draw,label=north:({\footnotesize -0.23016})](r) {$+$};
    \node[circle,draw](l11) [below left = of r,label=west:({\footnotesize -0.0672})] {$\times$};
    \node[circle,draw](l12) [below right = of r,label=west:({\footnotesize -0.30})] {$\times$};
    \node(l21) [below left = of l11,label=south:({\footnotesize 1})] {$X_1$};
    \node[circle,draw](l22) [below  = of l11,label=north:({\footnotesize -0.0672})] {$+$};
    \node[circle,draw](l23) [below = of l12,,label=north:({\footnotesize -0.30})] {$+$};
    \node(l24) [below right = of l12,label=south:({\footnotesize 1})] {$\neg X_1$};
    \node[circle,draw](l31) [below left = of l22,label=west:({\footnotesize -0.056})] {$\times$};
    \node[circle,draw](l32) [below left = of l23,label=west:({\footnotesize -0.084})] {$\times$};
    \node[circle,draw](l33) [below right = of l23,label=west:({\footnotesize -0.354})] {$\times$};
    \node[circle,draw](l41) [below left = of l31,label=north:({\footnotesize -0.4})] {$+$};
    \node(l42) [below  = of l31,label=south:({\footnotesize 1})] {$\neg X_2$};
    \node[circle,draw](l43) [right = of l42,label=north:({\footnotesize 0.14})] {$+$};
    \node(l44) [right = of l43,label=south:({\footnotesize 1})] {$X_2$};
    \node[circle,draw](l45) [right = of l44,label=north:({\footnotesize -0.6})] {$+$};
    \node(l46) [right = of l45,label=south:({\footnotesize 1})] {$\neg X_2$};
    \node[circle,draw](l47) [right = of l46,label=north:({\footnotesize 0.59})] {$+$};
    \node(l51) [below left = of l41,label=south:({\footnotesize 1})] {$X_3^c$};
    \node(l52) [right  = of l51,label=south:({\footnotesize -1})] {$\neg X_3^c$};
    \node(l53) [right = of l52,label=south:({\footnotesize 0.05})] {$X_4^c$};
    \node(l54) [right = of l53,label=south:({\footnotesize 0.95})] {$\neg X_4^c$};
    \node(l55) [right = of l54,label=south:({\footnotesize 1})] {$X_3^c$};
    \node(l56) [right = of l55,label=south:({\footnotesize -1})] {$\neg X_3^c$};
    \node(l57) [right = of l56,label=south:({\footnotesize 0.05})] {$X_4^c$};
    \node(l58) [right = of l57,label=south:({\footnotesize 0.95})] {$\neg X_4^c$};
    
    \draw[->] (r) edge node[left]{$0.3$} (l11);
    \draw[->] (r) edge node[right]{$0.7$}(l12);
    \draw[->] (l11) edge (l21);
    \draw[->] (l11) edge (l22);
    \draw[->] (l12) edge (l23);
    \draw[->] (l12) edge (l24);
    \draw[->] (l22) edge node[left]{$0.6$} (l31);
    \draw[->] (l22) edge node[right]{$0.4$} (l32);
    \draw[->] (l23) edge node[left]{$0.2$}(l32);
    \draw[->] (l23) edge node[right]{$0.8$} (l33);

    \draw[->] (l31) edge (l41);
    \draw[->] (l31) edge (l42);
    \draw[->] (l31) edge (l43);

    \draw[->] (l32) edge (l43);
    \draw[->] (l32) edge (l44);
    \draw[->] (l32) edge (l45);

    \draw[->] (l33) edge (l45);
    \draw[->] (l33) edge (l46);
    \draw[->] (l33) edge (l47);

    \draw[->] (l41) edge node[left]{$0.3$}(l51);
    \draw[->] (l41) edge node[right]{$0.7$}(l52);
    \draw[->] (l43) edge node[left]{$0.9$}(l53);
    \draw[->] (l43) edge node[right]{$0.1$}(l54);
    \draw[->] (l45) edge node[left]{$0.2$}(l55);
    \draw[->] (l45) edge node[right]{$0.8$}(l56);
    \draw[->] (l47) edge node[left]{$0.4$}(l57);
    \draw[->] (l47) edge node[right]{$0.6$}(l58);
    \eat{
    \node(r) at (0,6) {$+$};
    \node(l1) at (1,5){$\times$};
    \node(l1) at (1,7){$\times$};
    }
    \eat{
    \node (a) {$A$};
    \node (z) [right = of a] {$Z$};
    \node (b) [right = of z] {$B$};
    \node (x) [below left = of z] {$X$};
    \node (y) [below right = of z] {$Y$};
    \node (c) [below right = of x] {$C$};
    \path (a) edge (x);
    \path (a) edge (z);
    \path (b) edge (y);
    \path (b) edge (z);
    \path (c) edge (x);
    \path (c) edge (y);
    \path (x) edge (y);
    \path (z) edge (x);
    \path (z) edge (y); 
    }
    \end{tikzpicture}
}
    \caption{}
    \end{subfigure}
    \begin{subfigure}[b]{0.49\textwidth}
    \resizebox{0.95\textwidth}{!}{
    \begin{tikzpicture}[thick,scale=\myscale]
    \node[circle,draw,label=north:({\footnotesize 0.06355})](r) {$+$};
    \node[circle,draw](l11) [below left = of r,label=west:({\footnotesize 0.2118})] {$\times$};
    \node[circle,draw](l12) [below right = of r,label=west:({\footnotesize 0})] {$\times$};
    \node(l21) [below left = of l11,label=south:({\footnotesize 1})] {$X_1$};
    \node[circle,draw](l22) [below  = of l11,label=north:({\footnotesize 0.2118})] {$+$};
    \node[circle,draw](l23) [below = of l12,,label=north:({\footnotesize 0})] {$+$};
    \node(l24) [below right = of l12,label=south:({\footnotesize 1})] {$\neg X_1$};
    \node[circle,draw](l31) [below left = of l22,label=west:({\footnotesize 0.2432})] {$\times$};
    \node[circle,draw](l32) [below left = of l23,label=west:({\footnotesize 0.1648})] {$\times$};
    \node[circle,draw](l33) [below right = of l23,label=west:({\footnotesize -0.0412})] {$\times$};
    \node[circle,draw](l41) [below left = of l31,label=north:({\footnotesize 0.304})] {$+$};
    \node(l42) [below  = of l31,label=south:({\footnotesize 1})] {$\neg X_2$};
    \node[circle,draw](l43) [right = of l42,label=north:({\footnotesize 0.8})] {$+$};
    \node(l44) [right = of l43,label=south:({\footnotesize 1})] {$X_2$};
    \node[circle,draw](l45) [right = of l44,label=north:({\footnotesize 0.206})] {$+$};
    \node(l46) [right = of l45,label=south:({\footnotesize 1})] {$\neg X_2$};
    \node[circle,draw](l47) [right = of l46,label=north:({\footnotesize -0.2})] {$+$};
    \node(l51) [below left = of l41,label=south:({\footnotesize 0.99})] {$X_3^c$};
    \node(l52) [right  = of l51,label=south:({\footnotesize 0.01})] {$\neg X_3^c$};
    \node(l53) [right = of l52,label=south:({\footnotesize 1})] {$X_4^c$};
    \node(l54) [right = of l53,label=south:({\footnotesize -1})] {$\neg X_4^c$};
    \node(l55) [right = of l54,label=south:({\footnotesize 0.99})] {$X_3^c$};
    \node(l56) [right = of l55,label=south:({\footnotesize 0.01})] {$\neg X_3^c$};
    \node(l57) [right = of l56,label=south:({\footnotesize 1})] {$X_4^c$};
    \node(l58) [right = of l57,label=south:({\footnotesize -1})] {$\neg X_4^c$};
    
    \draw[->] (r) edge node[left]{$0.3$} (l11);
    \draw[->] (r) edge node[right]{$0.7$}(l12);
    \draw[->] (l11) edge (l21);
    \draw[->] (l11) edge (l22);
    \draw[->] (l12) edge (l23);
    \draw[->] (l12) edge (l24);
    \draw[->] (l22) edge node[left]{$0.6$} (l31);
    \draw[->] (l22) edge node[right]{$0.4$} (l32);
    \draw[->] (l23) edge node[left]{$0.2$}(l32);
    \draw[->] (l23) edge node[right]{$0.8$} (l33);

    \draw[->] (l31) edge (l41);
    \draw[->] (l31) edge (l42);
    \draw[->] (l31) edge (l43);

    \draw[->] (l32) edge (l43);
    \draw[->] (l32) edge (l44);
    \draw[->] (l32) edge (l45);

    \draw[->] (l33) edge (l45);
    \draw[->] (l33) edge (l46);
    \draw[->] (l33) edge (l47);

    \draw[->] (l41) edge node[left]{$0.3$}(l51);
    \draw[->] (l41) edge node[right]{$0.7$}(l52);
    \draw[->] (l43) edge node[left]{$0.9$}(l53);
    \draw[->] (l43) edge node[right]{$0.1$}(l54);
    \draw[->] (l45) edge node[left]{$0.2$}(l55);
    \draw[->] (l45) edge node[right]{$0.8$}(l56);
    \draw[->] (l47) edge node[left]{$0.4$}(l57);
    \draw[->] (l47) edge node[right]{$0.6$}(l58);
    \eat{
    \node(r) at (0,6) {$+$};
    \node(l1) at (1,5){$\times$};
    \node(l1) at (1,7){$\times$};
    }
    \eat{
    \node (a) {$A$};
    \node (z) [right = of a] {$Z$};
    \node (b) [right = of z] {$B$};
    \node (x) [below left = of z] {$X$};
    \node (y) [below right = of z] {$Y$};
    \node (c) [below right = of x] {$C$};
    \path (a) edge (x);
    \path (a) edge (z);
    \path (b) edge (y);
    \path (b) edge (z);
    \path (c) edge (x);
    \path (c) edge (y);
    \path (x) edge (y);
    \path (z) edge (x);
    \path (z) edge (y); 
    }
    \end{tikzpicture}
}
    \caption{}
    \end{subfigure}
    \caption{(a) Value computations for partial derivative of the QPC given in Figure 1 in the main paper w.r.t. $X_3^c$ and (b) Value computations for partial derivative of the QPC given in Figure 1 in the main paper w.r.t. $X_4^c$. The values of the leaf, sum and product nodes are given in brackets on their bottom, top and left respectively. The value of the root node equals the partial derivative.}
    \label{fig:qpc-derivatives}
\end{figure*}

\begin{example}
    Figures \ref{fig:qpc-derivatives}(a) and (b) 
 show the value computations for the partial derivative of the QPC w.r.t. $X_3^c$ and $X_4^c$ respectively.  
\end{example}

\section{Experimental Setup and Details}

\subsection{An Overview of the Datasets and Models}

We furnish comprehensive information regarding the datasets employed for conducting our experiments in Tables \ref{tab:dataset-details} and \ref{tab:dataset-details-comp}, along with the PC models utilized for these evaluations. Our selection of datasets is deliberate, aiming to encompass a diverse range of scenarios and serve as representative instances to assess the performance and scalability of our algorithms comprehensively. It's important to note that for the EMNIST dataset, we select classes exclusively from the EMNIST Letters subset.
\begin{table}[ht!]
\centering
\resizebox{0.5\columnwidth}{!}{%
\begin{tabular}{@{}ccc@{}}
\toprule
\textbf{Dataset} & \textbf{Number of Variables} & \textbf{Number of Nodes in PC} \\ \midrule
nltcs            & 16                           & 125                             \\
msnbc            & 17                           & 46                              \\
kdd              & 64                           & 274                             \\
plants           & 69                           & 3737                            \\
baudio           & 100                          & 348                             \\
bnetflix         & 100                          & 400                             \\
jester           & 100                          & 274                             \\
accidents        & 111                          & 1178                            \\
mushrooms        & 112                          & 902                             \\
connect4         & 126                          & 2128                            \\
tretail          & 135                          & 359                             \\
rcv1             & 150                          & 519                             \\
pumsb star       & 163                          & 2860                            \\
dna              & 180                          & 1855                            \\
kosarek          & 190                          & 779                             \\
tmovie           & 500                          & 7343                            \\
book             & 500                          & 1628                            \\
cwebkb           & 839                          & 3154                            \\
cr52             & 889                          & 7348                            \\
c20ng            & 910                          & 2467                            \\
moviereview      & 1001                         & 2567                            \\
bbc              & 1058                         & 3399                            \\ \bottomrule
\end{tabular}%
}
\caption{Overview of Dataset and PC Model Descriptions for TPM Datasets}
\label{tab:dataset-details}
\end{table}

\begin{table}[ht!]
\centering
\resizebox{0.5\columnwidth}{!}{%
\begin{tabular}{@{}cccc@{}}
\toprule
\textbf{Dataset Name} & \textbf{Class} & \textbf{Number of Variables} & \textbf{Number of Nodes in PC} \\ \midrule
cifar10               & 0              & 512                          & 1294                   \\
cifar10               & 1              & 512                          & 1326                   \\
cifar10               & 2              & 512                          & 1280                   \\
cifar10               & 3              & 512                          & 1684                   \\
cifar10               & 4              & 512                          & 1743                   \\
cifar10               & 5              & 512                          & 1632                   \\
cifar10               & 6              & 512                          & 1796                   \\
cifar10               & 7              & 512                          & 1359                   \\
cifar10               & 8              & 512                          & 1300                   \\
cifar10               & 9              & 512                          & 1279                   \\ \midrule
emnist                & 1              & 784                          & 2125                   \\
emnist                & 2              & 784                          & 2124                   \\
emnist                & 3              & 784                          & 2100                   \\
emnist                & 4              & 784                          & 2122                   \\
emnist                & 19             & 784                          & 2684                   \\
emnist                & 20             & 784                          & 2147                   \\
emnist                & 22             & 784                          & 2060                   \\
emnist                & 24             & 784                          & 2734                   \\
emnist                & 26             & 784                          & 2137                   \\ \midrule
mnist                 & 0              & 784                          & 3960                   \\
mnist                 & 1              & 784                          & 4324                   \\
mnist                 & 2              & 784                          & 4660                   \\
mnist                 & 3              & 784                          & 4465                   \\
mnist                 & 4              & 784                          & 4510                   \\
mnist                 & 5              & 784                          & 4063                   \\
mnist                 & 6              & 784                          & 3776                   \\
mnist                 & 7              & 784                          & 4408                   \\
mnist                 & 8              & 784                          & 3804                   \\
mnist                 & 9              & 784                          & 3714                   \\ \bottomrule
\end{tabular}%
}
\caption{CIFAR, MNIST, EMNIST Letters: Dataset and PC Overview}
\label{tab:dataset-details-comp}
\end{table}

\subsection{Hyperparameter Selection}
In our experimental setup, we maintained a consistent minibatch size of 128 instances across all conducted experiments. To ensure effective training, we adopted a learning rate decay strategy wherein the learning rate was reduced by a factor of 0.9 after a fixed number of training epochs. As detailed in the primary text, the selection of optimal hyperparameters was accomplished through 5-fold cross validation.

In the context of discrete scenarios, a single hyperparameter $\alpha$ was required. To determine the most suitable value for $\alpha$, we conducted an exploration within the range of $\{0.01, 0.1, 1, 10, 100, 1000\}$. This systematic approach to hyperparameter tuning contributes to the reliability and generalizability of our experimental results.

\subsection{Label Generation for Supervised Learning Benchmarks}
\label{sec:sup_label_gen}
In the process of comparing the performance of \our method with the supervised learning approach outlined in the paper, we adopted the Stochastic Hill Climbing Search. This selection was guided by the challenge posed by the substantial number of potential query variables present in our experimental setup. Given the complexity of obtaining exact solutions within such a context, we opted for an alternative method that aligns with practical feasibility.

Our approach involved initializing the query variables randomly and conducting 1000 iterations for each specific problem instance. This iterative process yielded labels that were subsequently employed to train a supervised neural network. By employing this methodology, we aimed to effectively navigate the challenges posed by a significant number of query variables, facilitating a meaningful comparative evaluation between \our and the supervised learning approach.

\section{Inference Time Comparison on TPM Datasets}
\begin{figure}[h!]
    \centering
    \begin{subfigure}[b]{0.49\textwidth}
        \centering
        \includegraphics[width=\linewidth]{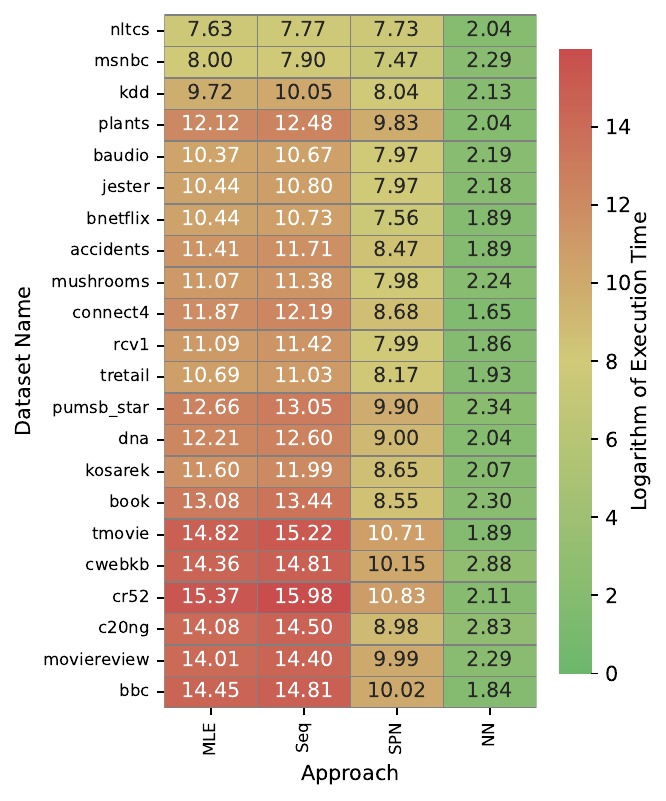}
        \caption{\mpe}
        \label{fig:mpe_time}
    \end{subfigure}
    \hfill 
    \begin{subfigure}[b]{0.49\textwidth}
        \centering
        \includegraphics[width=\linewidth]{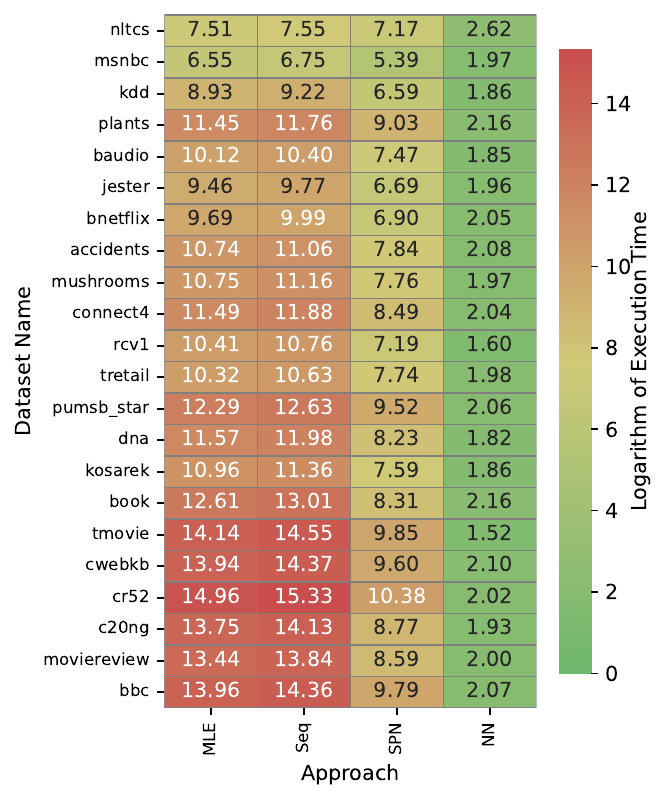}
        \caption{\mmap}
        \label{fig:mmap_time}
    \end{subfigure}
    \caption{Heatmap illustrating inference time for \mle, \seq, \spn, and \our methods on a Logarithmic micro-second scale. The color green indicates shorter (more favorable) time.}
    \label{fig:time_for_methods}
\end{figure}

We have presented the inference times for both the polytime baseline methods and our proposed \our method, as illustrated in Figure \ref{fig:time_for_methods}. The inference times for MPE queries are depicted in Figure \ref{fig:mpe_time}, while those for MMAP queries are showcased in Figure \ref{fig:mmap_time}. The values in each cell correspond to the natural logarithm of the time in microseconds for each method and dataset. Lower values are represented by green, while higher values are denoted by red. Notably, our method exhibits a substantial performance advantage over all other methods. Additionally, it is noteworthy that as the dataset size (and consequently the SPN size) increases, the inference time for the baseline methods (\mle, \seq, and \spn) increases, whereas our method's inference time remains almost the same. This distinction arises because the baseline methods depend on the SPN size during inference, whereas our approach allows us to regulate inference time by modifying the neural network's size without impacting the SPN.

\section{A Comparative Analysis: \our Method Versus Supervised Learning Approaches}
\begin{figure}[ht!]
    \centering
    \begin{subfigure}[b]{0.45\textwidth}
        \centering
        \includegraphics[width=\linewidth]{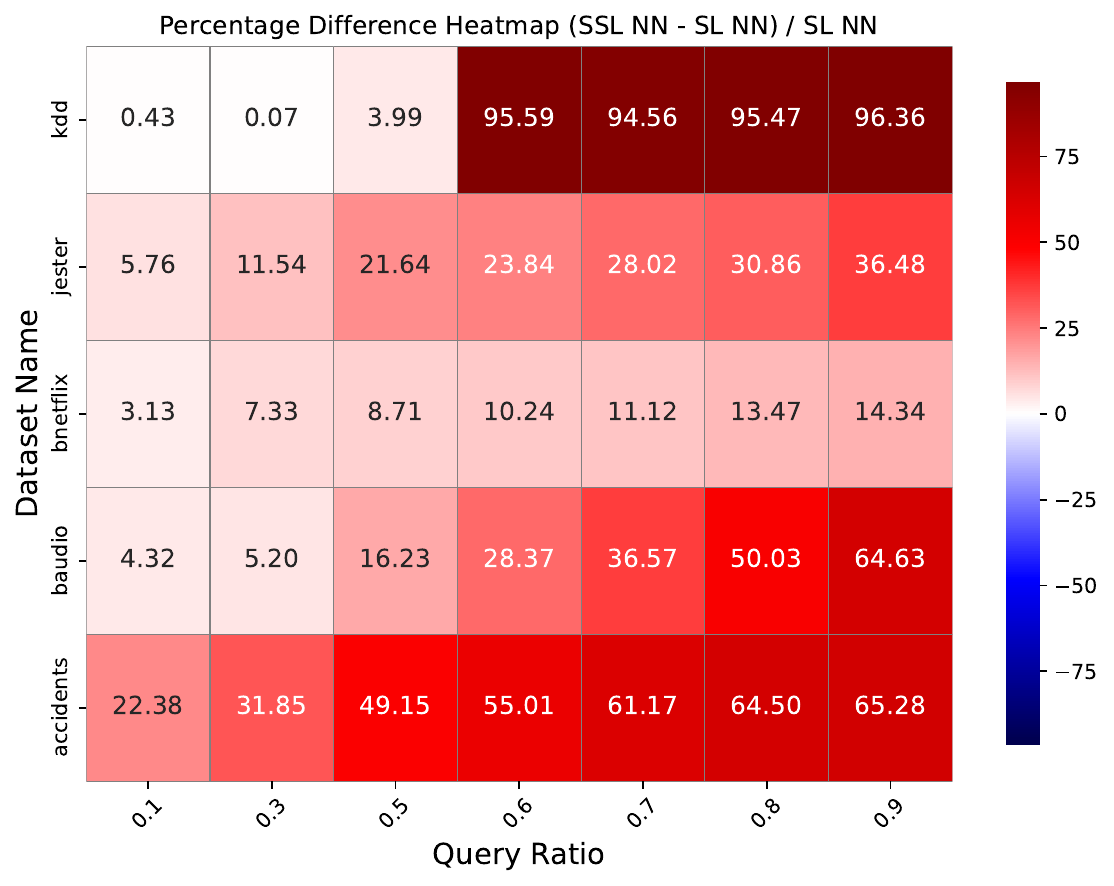}
        \caption{\mpe}
        \label{fig:mpe_sl_vs_ssl}
    \end{subfigure}
    \hfill 
    \begin{subfigure}[b]{0.45\textwidth}
        \centering
        \includegraphics[width=\linewidth]{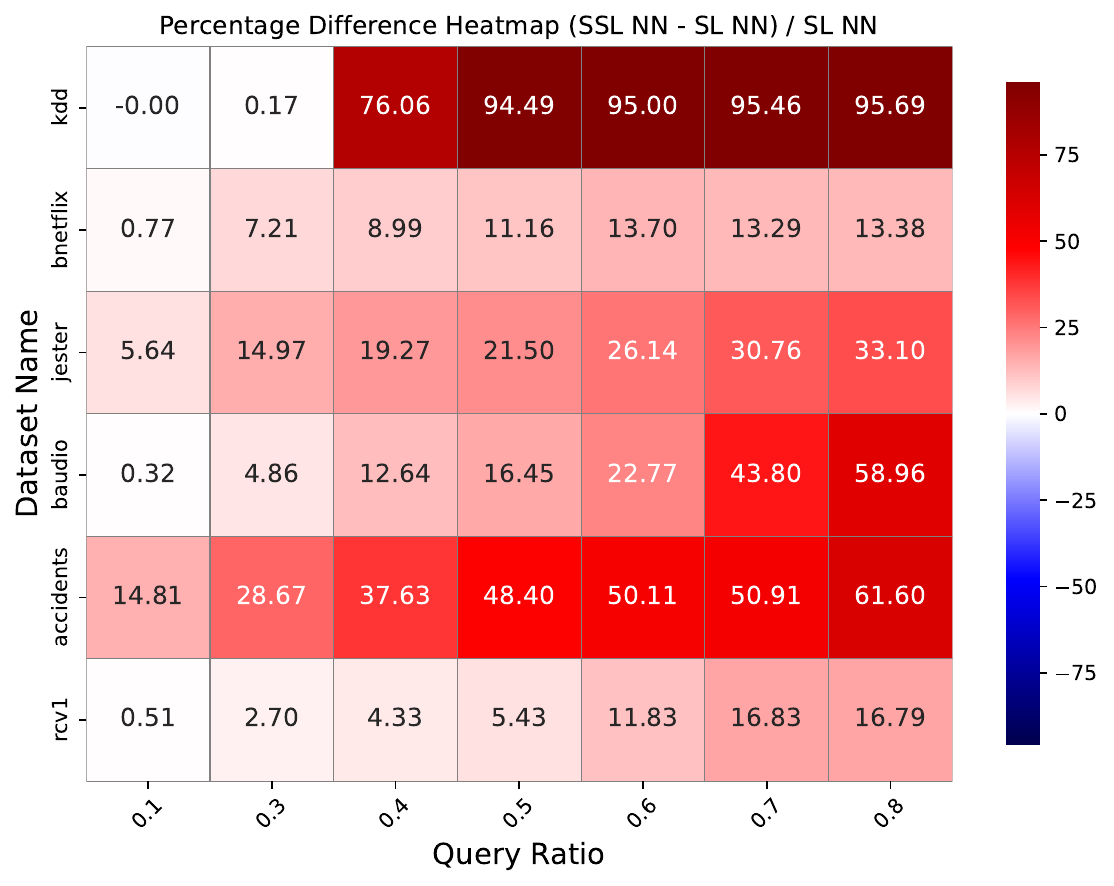}
        \caption{\mmap}
        \label{fig:mmap_sl_vs_ssl}
    \end{subfigure}
    \caption{Heatmap showing the percentage difference in log-likelihood scores between SSMP and Supervised Learning method. Blue color represents the supervised method's superiority (negative values), while red color represents \our's superiority (positive values). The datasets are arranged in ascending order of their number of variables.}
    \label{fig:sl_vs_ssl}
\end{figure}

We conduct a comparative analysis of our approach against supervised learning methods to assess its performance in the heatmaps shown in Figure \ref{fig:sl_vs_ssl}. The heatmap on the left illustrates the percentage difference in log-likelihood scores for MPE, while the one on the right depicts the corresponding difference for MMAP. Labels for training the supervised method are obtained through the procedure outlined in the previous section. Subsequently, training is executed using the mean squared error (MSE) criterion. We adopt an identical neural network architecture as employed in \our method, maintaining consistent training protocols across all aspects except for the training procedure and loss function. This facilitates a direct comparison of these methods, exclusively in terms of their training processes and associated losses. It's worth noting that generating accurate labels in this context presents a significant challenge due to the substantial volume of query variables involved.

By observing the heatmaps, we can confirm that our method consistently surpasses the performance of the supervised approach across all cases, except for one instance (specifically, for the kdd dataset and query = 0.1 for MMAP), where both methods exhibit equivalent log-likelihood scores. The supervised method closely aligns with our approach for smaller query sets across most datasets. However, as the query variable count grows, the supervised method's efficacy diminishes. This trend is also pronounced when increasing the variable count within the datasets, leading to a decrease in the supervised method's performance, as depicted in the heatmap. 
This underscores the need for a self-supervised approach that operates independently of true labels. Notably, the training duration for supervised methods encompasses the time required to obtain the true labels, whereas our method depends solely on a trained probabilistic classifier (PC) to propagate the loss.

\section{Negative Log Likelihood Scores: \our Method vs. Baseline Methods}
We present the negative log-likelihood scores for all datasets in Figures \ref{fig:mpeNLTCS} through \ref{fig:mpebbc} for the MPE method, and in Figures \ref{fig:mmapNLTCS} through \ref{fig:mmapbbc} for the MMAP method. Each bar corresponds to the mean value of the respective method, while the tick marks denote the mean ± standard deviation. Note that since we are considering the negative of the log likelihood means, lower values indicate superior performance by the method. These detailed visualizations enable us to grasp the performance characteristics of our method, alongside the baseline techniques including \spn, \mle, and \seq, across diverse datasets, query ratios, and evidence ratios. Across all the plots, negative log likelihood scores for \spn are depicted using blue bars, \our method's scores are indicated by yellow bars, green bars represent scores for \mle method, and \seq method's scores are visualized through red bars.
\newcommand{\datasetnames}{%
  NLTCS, MSNBC, kdd, plants, baudio, jester, bnetflix, accidents, mushrooms, connect4, rcv1, tretail, pumsb, dna, kosarek, book, tmovie, cwebkb, cr52, c20ng, moviereview, bbc}

\subsection{Scores for \mpe}

\foreach \dataset in \datasetnames {%
  \begin{figure}[htp!]
    \centering
    \includegraphics[width=\columnwidth]{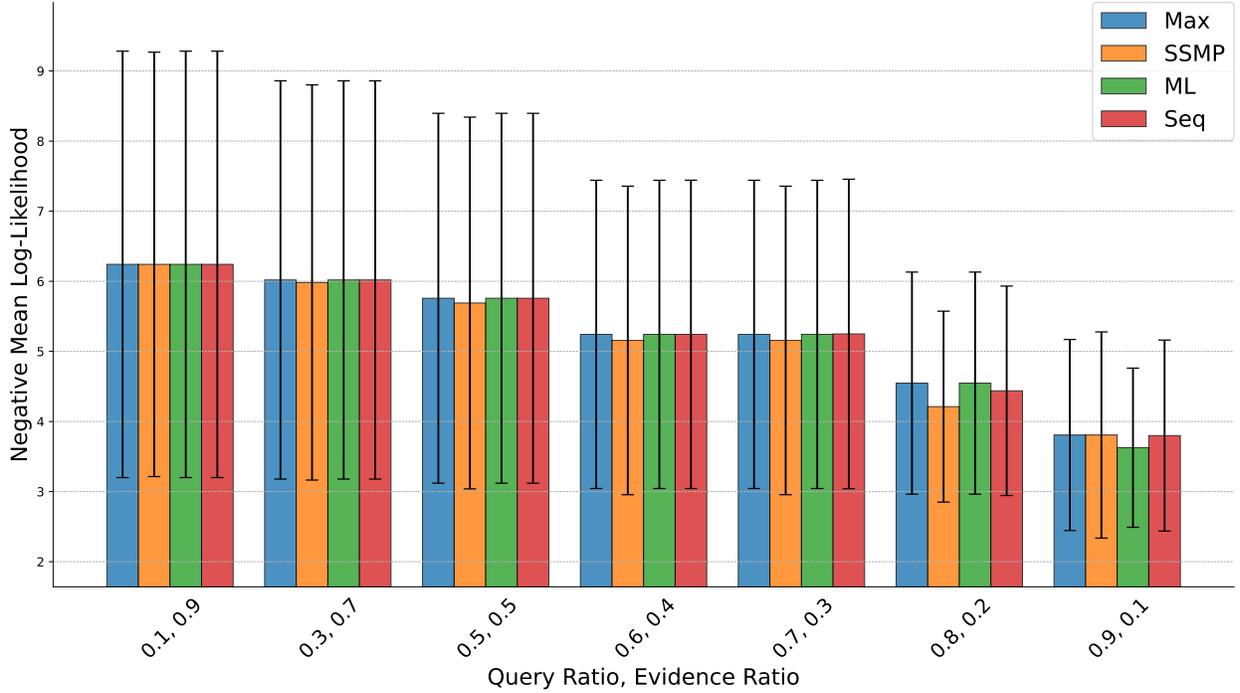}
    \caption{Negative Log-Likelihood Scores for \our Method and Baselines on \dataset \xspace for \mpe. Lower Scores Indicate Better Performance}
    \label{fig:mpe\dataset}
  \end{figure}
}

Figures \ref{fig:mpeNLTCS} to \ref{fig:mpebbc} display the results of the \mpe task on the TPM datasets. As indicated by the heatmaps and contingency tables in the main paper, our method performs well compared to other baselines. Notably, as the query set size increases, our method's improvement becomes more pronounced. For smaller query ratios, our method competes reasonably well with other methods, with only a few datasets where it falls slightly short. 

\subsection{Scores for \mmap}
\foreach \dataset in \datasetnames {%
  \begin{figure}[htp!]
    \centering
    \includegraphics[width=\columnwidth]{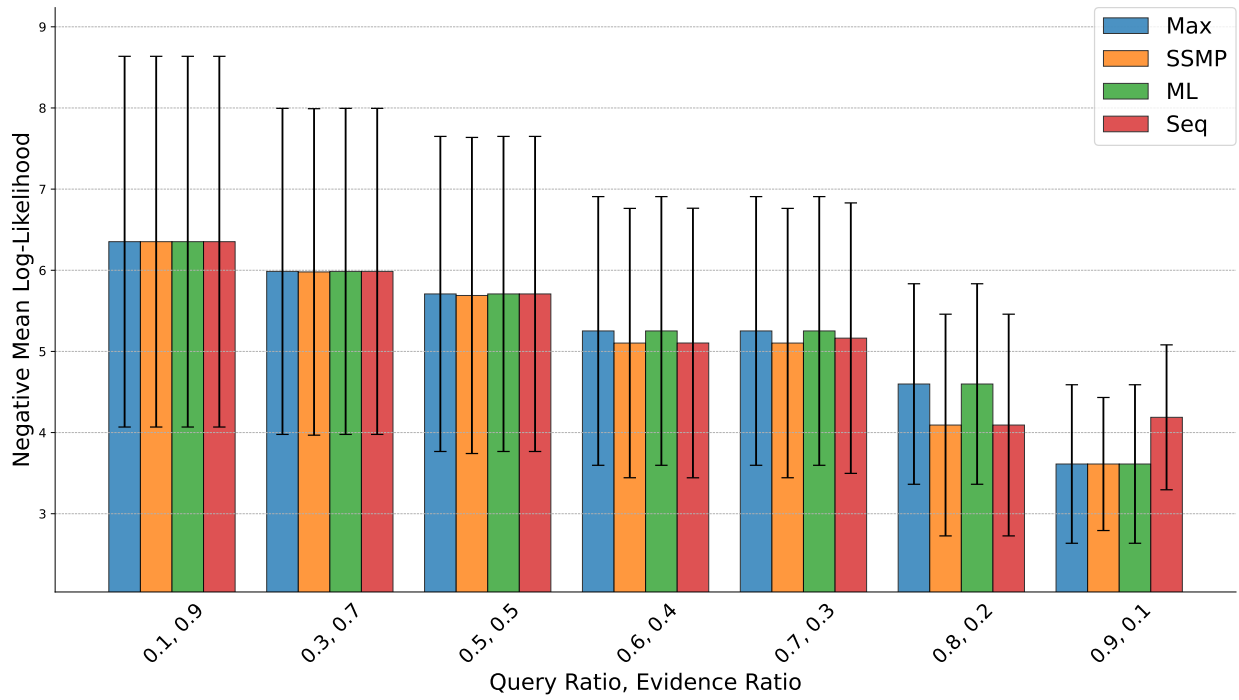}
    \caption{Negative Log-Likelihood Scores for \our Method and Baselines on \dataset \xspace for \mmap. Lower Scores Indicate Better Performance.}
    \label{fig:mmap\dataset}
  \end{figure}
}

Similar patterns emerge when examining figures \ref{fig:mmapNLTCS} through \ref{fig:mmapbbc}, which pertain to the \mmap task. Notably, \our's performance excels even further in this task. Interestingly, it also demonstrates a slight advantage for smaller queries in comparison to its performance in the \mpe task. This underscores a distinct attribute of our method: it's capacity to enhance its performance as task complexity increases, a quality that sets it apart from other approaches. 

Our method excels in the \mpe task on TPM datasets, outperforming baselines for larger query sets. Notably, it exhibits adaptive prowess in the \mmap task, showcasing improved performance for smaller queries. 
\newpage

\section{Completing Images in MNIST and EMNIST: Enhancing Visual Coherence}
\begin{figure}[ht!]
    \centering
    \setlength{\tabcolsep}{10pt} 
    
    \begin{tabular}{@{}cc@{}}
        \begin{subfigure}{0.42\columnwidth}
            \includegraphics[width=\linewidth]{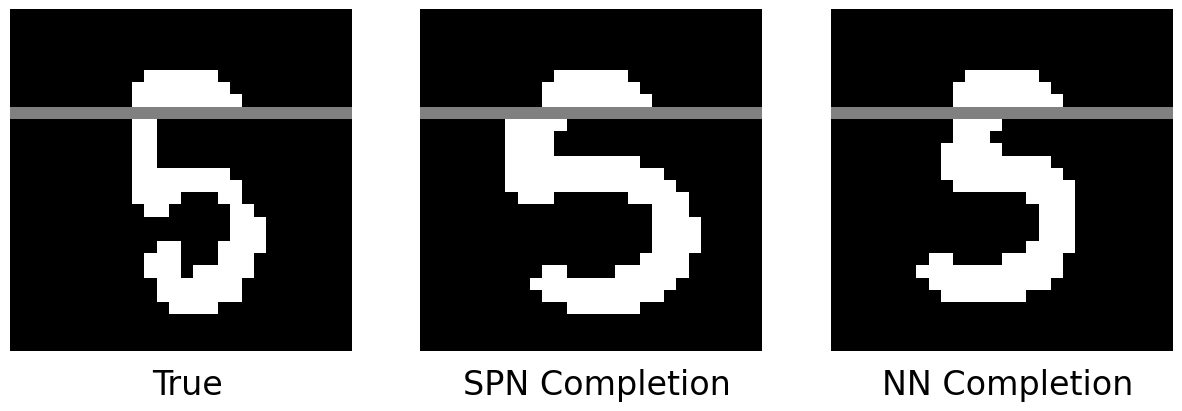}
            \caption{MNIST Digit 5}
        \end{subfigure} &
        
        \begin{subfigure}{0.42\columnwidth}
            \includegraphics[width=\linewidth]{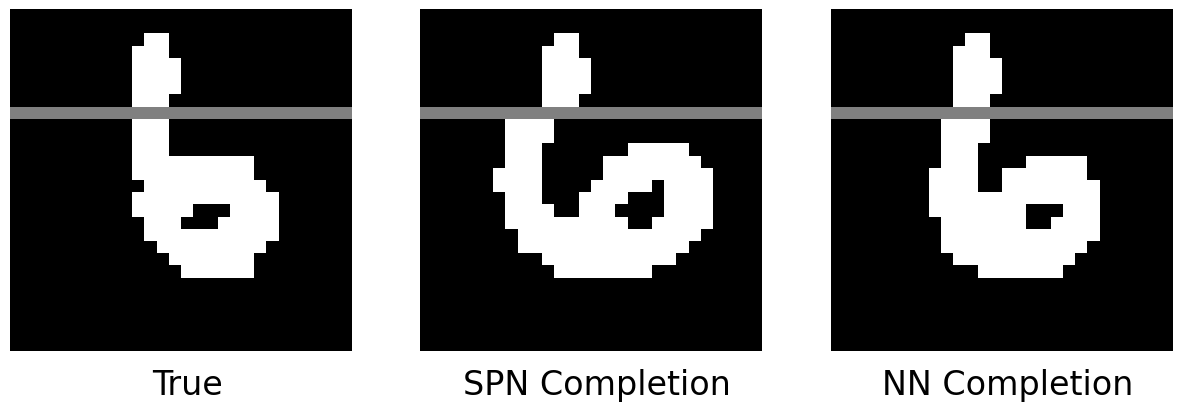}
            \caption{MNIST Digit 6}
        \end{subfigure} \\
        
        \begin{subfigure}{0.42\columnwidth}
            \includegraphics[width=\linewidth]{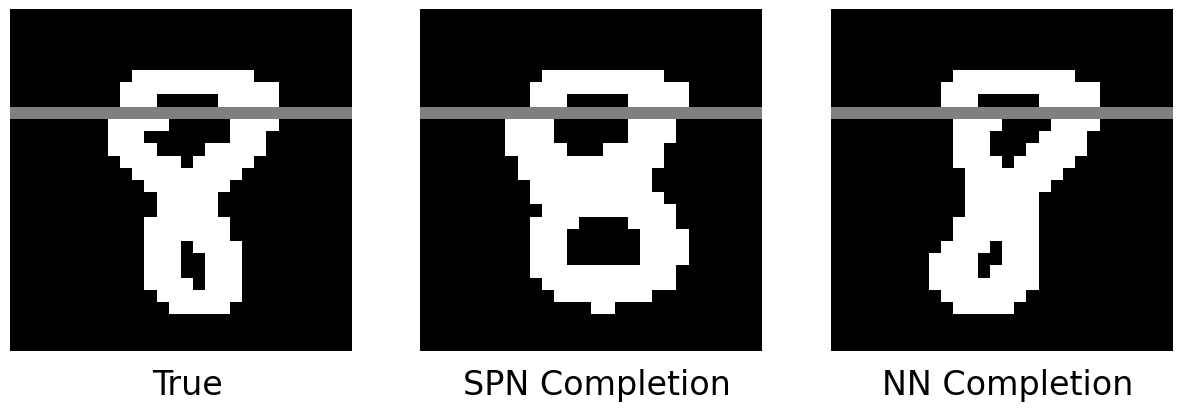}
            \caption{MNIST Digit 8}
        \end{subfigure} &
        
        \begin{subfigure}{0.42\columnwidth}
            \includegraphics[width=\linewidth]{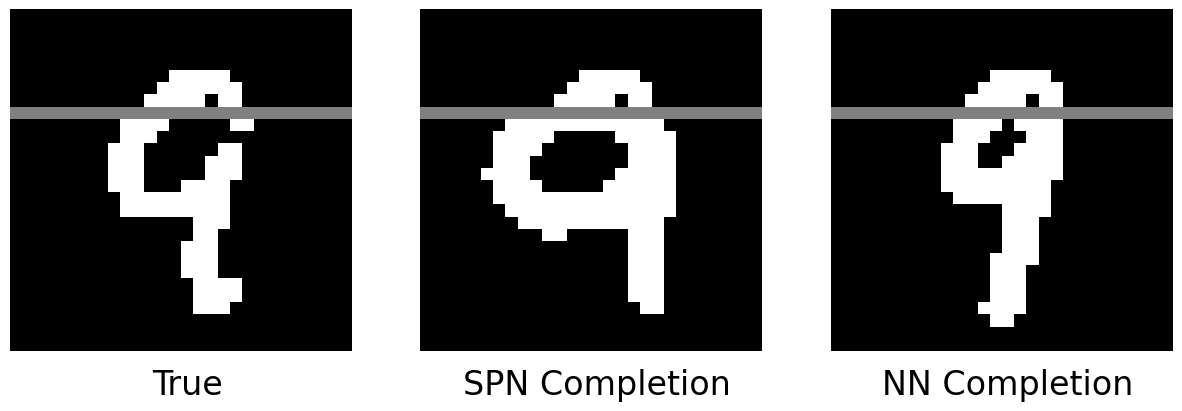}
            \caption{MNIST Digit 9}
        \end{subfigure} \\
    \end{tabular}
    
    \vspace{1em} 
    
    \begin{tabular}{@{}cc@{}}
        \begin{subfigure}{0.42\columnwidth}
            \includegraphics[width=\linewidth]{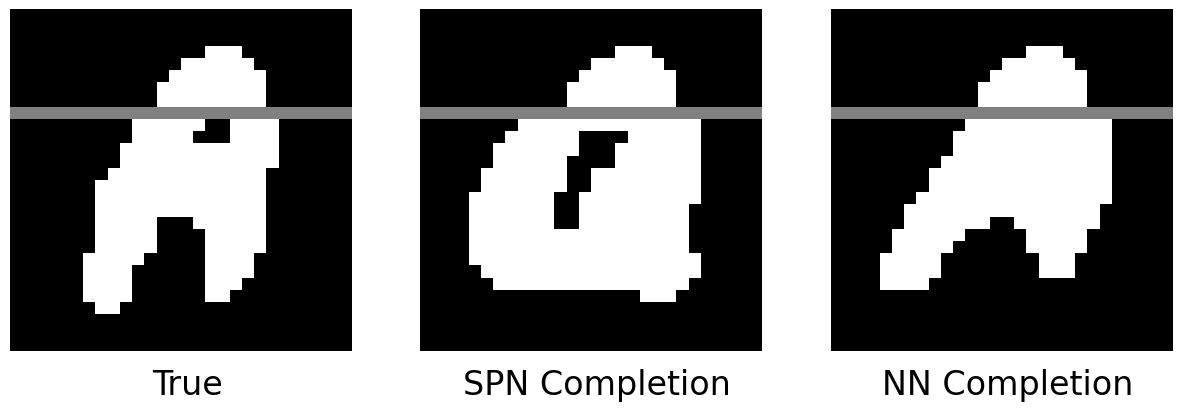}
            \caption{EMNIST Character A}
        \end{subfigure} &
        
        \begin{subfigure}{0.42\columnwidth}
            \includegraphics[width=\linewidth]{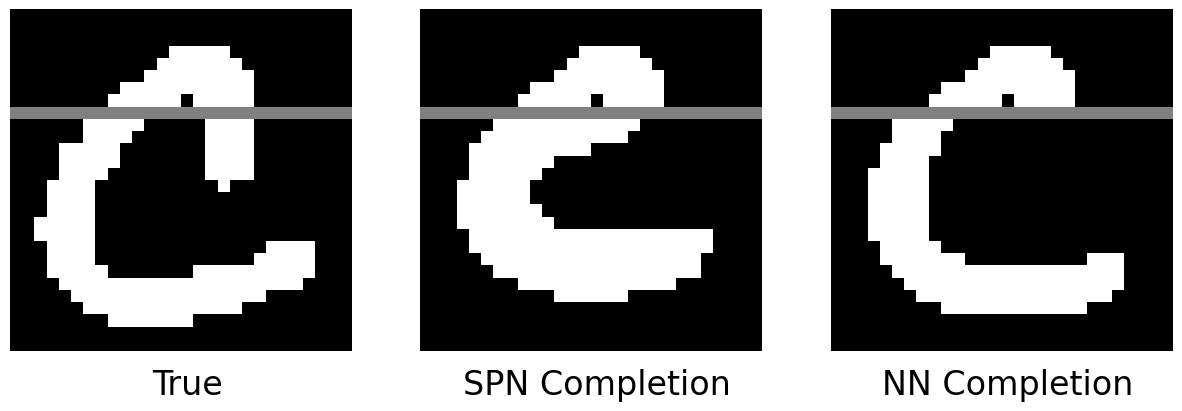}
            \caption{EMNIST Character C}
        \end{subfigure} \\
        
        \begin{subfigure}{0.42\columnwidth}
            \includegraphics[width=\linewidth]{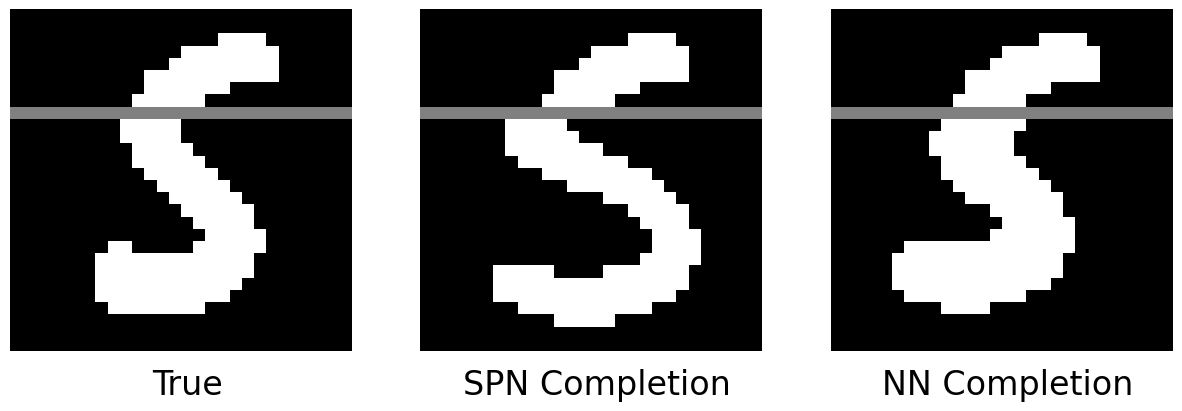}
            \caption{EMNIST Character S}
        \end{subfigure} &
        
        \begin{subfigure}{0.42\columnwidth}
            \includegraphics[width=\linewidth]{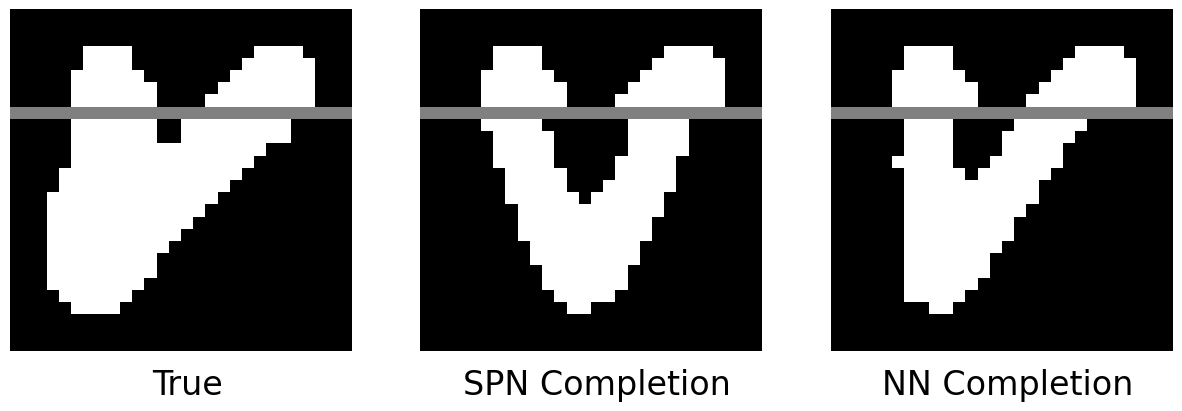}
            \caption{EMNIST Character V}
        \end{subfigure} \\
    \end{tabular}
    
    \caption{Image Completions for MNIST and EMNIST datasets. First image: Original. Second image: \spn Completion. Third image: \our Completion. The gray line separates evidence (above) from query (below).}
    \label{fig:image-completion}
\end{figure}

Qualitative image completions for MNIST and EMNIST datasets are presented in Figure \ref{fig:image-completion} for MNIST Digits \{5, 6, 8, 9\} and EMNIST characters \{a, c, s, v\} by \our and \spn. Beginning with the original image as the initial reference, followed by the \spn method's completion attempt, and culminating in our novel \our method's completion. Notably, \our demonstrates equal or superior performance compared to the \spn method. In particular, characters such as 8, 9, A, and S display discrepancies in their \spn-generated completions, where the lower segments fail to seamlessly align with their upper counterparts. In stark contrast, the proposed \our method offers a remarkable improvement in completion quality. It adeptly addresses the difficulties posed by limited available evidence and generates reconstructions that are not only smoother but also exhibit remarkable coherence. This highlights \our method's efficacy in real-world scenarios with limited information.

\end{document}